\documentclass[5p,authoryear]{elsarticle} 

\usepackage{framed,multirow}
\usepackage{amssymb}
\usepackage{latexsym}
\usepackage{xcolor}
\usepackage{lineno,hyperref}
\modulolinenumbers[5]
\usepackage{amsmath}
\usepackage{multirow}
\usepackage{color,graphicx}
\usepackage{nicematrix}
\usepackage{arydshln}
\usepackage{hyperref}
\usepackage{url}
\usepackage{footnote}
\usepackage{bm}
\usepackage{cite}
\usepackage{subfigure}
\usepackage{verbatim}
\usepackage{ulem}
\usepackage{graphicx}
\usepackage{pgfplots}
\usepackage{cancel}

\usepackage{booktabs}  
\usepackage{tabularx}

\usepackage{enumitem}

\journal{Medical Image Analysis}

\begin{document}

\begin{frontmatter}

\title{Spectral Rectification for Parameter-Efficient Adaptation of Foundation Models in Colonoscopy Depth Estimation}

\author[label1,label2]{Xiaoxian Zhang\fnref{equal}} 
\author[label1]{Minghai Shi\fnref{equal}} 
\author[label1]{Lei Li\corref{cor1}} 

\address[label1]{Department of Biomedical Engineering, National University of Singapore, Singapore}
\address[label2]{Department of Radiotherapy, The First Affiliated Hospital of Xi'an Jiaotong University, Xi'an, China}

\cortext[cor1]{Corresponding author: Lei Li (\href{mailto:lei.li@nus.edu.sg}{lei.li@nus.edu.sg})}
\fntext[equal]{These authors contributed equally to this work and are co-first authors.}

\begin{abstract}
Accurate monocular depth estimation is critical in colonoscopy for lesion localization and navigation. Foundation models trained on natural images fail to generalize directly to colonoscopy. We identify the core issue not as a semantic gap, but as a statistical shift in the frequency domain: colonoscopy images lack the strong high-frequency edge and texture gradients that these models rely on for geometric reasoning. To address this, we propose SpecDepth, a parameter-efficient adaptation framework that preserves the robust geometric representations of the pre-trained models while adapting to the colonoscopy domain. Its key innovation is an adaptive spectral rectification module, which uses a learnable wavelet decomposition to explicitly model and amplify the attenuated high-frequency components in feature maps. Different from conventional fine-tuning that risks distorting high-level semantic features, this targeted, low-level adjustment realigns the input signal with the original inductive bias of the foundational model. On the public C3VD and SimCol3D datasets, SpecDepth achieved state-of-the-art performance with an absolute relative error of 0.022 and 0.027, respectively. Our work demonstrates that directly addressing spectral mismatches is a highly effective strategy for adapting vision foundation models to specialized medical imaging tasks. The code will be released publicly after the manuscript is accepted for publication.
\end{abstract}


\end{frontmatter}


\section{Introduction}


Colorectal cancer (CRC) represents a major and increasing global health burden, with an estimated 1.9 million new cases and 935, 000 deaths annually \citep{Sung2021}.
Beyond absolute numbers, the rising incidence of early-onset CRC in younger populations underscores its growing public health significance \citep{Xi2021}.
Colonoscopy remains gold standard for CRC screening, diagnosis, and therapy \citep{journal/Medicine/Cianci2024}.
However, its performance is highly operator-dependent and constrained by intrinsic limitations such as restricted 3D perception, incomplete mucosal coverage, specular highlights, and fluid-related artifacts \citep{Leufkens2012}.
Monocular depth estimation (MDE), which can reconstruct dense depth map from single RGB frames, offers a promising pathway to enhance 3D understanding in colonoscopy imaging \citep{journal/NC/ming2021}.
Nonetheless, achieving reliable MDE in colonoscopy remains challenging in practice \citep{journal/NC/ming2021,conf/NeurIPS/Yang2024}. 

Traditional MDE methods rely on restrictive depth cues, such as the shallow depth of field required in shape-from-focus or defocus techniques \citep{journal/TPAMI/Arampatzakis2023}.
Deep learning has recently emerged as a powerful alternative, showing strong potential to address the ill-posed nature of monocular depth prediction \citep{conf/CVPR/Bhat2021,journal/NC/ming2021}. 
The field have progressed from multi-scale CNNs \citep{Eigen2014} to attention-based designs \citep{journal/IJMLC/Chen2021}, with the modeling focus shifting from local appearance cues toward scene-level geometry \citep{conf/IPMI/Wang2023,journal/TPAMI/Shao2024}.
However, these methods typically require large annotated datasets \citep{Martyniak2025} and often learn pixel-to-depth mappings directly from RGB images \citep{journal/Sensors/Masoumian2022}.
As a result, the estimated depth maps are sensitive to illumination changes and scale ambiguities, while still lacking stable geometric structure \citep{journal/MedIA/Rau2024,Jong2025}.

Foundation models represent a transformative advancement beyond conventional deep learning paradigms, offering particularly promising potential for monocular depth estimation \citep{conf/ICCV/Caron2021,conf/ICCV/Wang2023}. 
Pre-trained on massive and diverse datasets, they provide powerful, transferable priors that can drastically reduce the annotation burden and offer stable performance under variable clinical conditions \citep{conf/ECCVW/SheikhZeinoddin2025}. 
A premier example in this domain is Depth Anything \citep{Yang2024cvpr} and Depth Anything V2 \citep{conf/NeurIPS/Yang2024}, which leverage scalable architectures and monumental training data to achieve remarkable generalization across various domains and tasks. 
Their robust pretrained weights and modular design make them exceptionally attractive candidates for transfer learning to specialized domains \citep{Wu2025, Zhao2023, conf/MICCAI/Cui2024}.
However, the deployment of foundation models in colonoscopy presents non-trivial challenges compared to structured modalities like CT and MRI where anatomical boundaries provide strong inductive biases \citep{Wu2025, Zhang2023}.
Current adaptation methodologies predominantly attribute this performance degradation to high-level semantic discrepancies, postulating that the visual semantics of the gastrointestinal tract diverge fundamentally from natural scenes \citep{journal/Access/Peng2024}.
Consequently, prevailing strategies employ intensive fine-tuning or extensive reparameterization to align the model with the target domain. 
Nevertheless, this approach introduces a fundamental trade-off: inducing sufficient plasticity to capture medical semantics risks catastrophic forgetting, effectively degrading the robust general-purpose priors that motivated the adoption of the foundation model. 
This suggests that characterizing the domain gap solely as a semantic issue is insufficient, and that such high-level adaptation may overlook the fundamental signal-level barriers limiting model transferability.

\begin{figure*}[t]\center
 \includegraphics[width=1\textwidth]{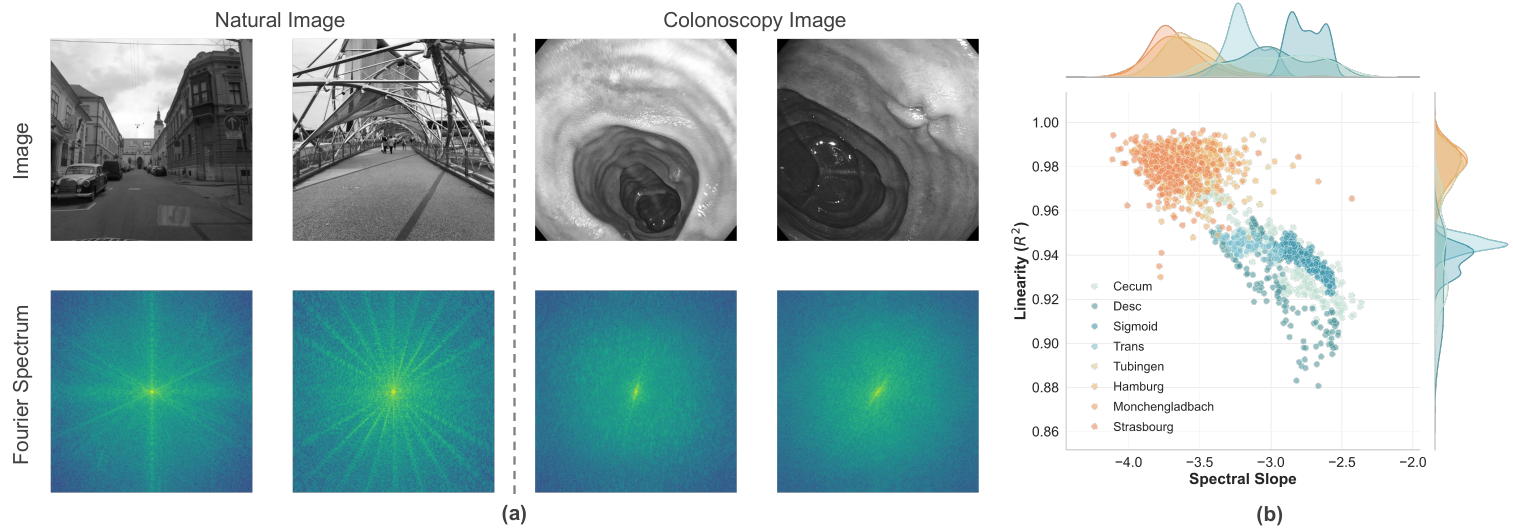}\\[-2ex]
   \caption{Statistical shift in the frequency domain between natural and colonoscopy images. 
   (a) Visual comparison of Fourier magnitude spectra. 
   Natural images (left) display prominent star-like anisotropic radiation patterns, reflecting the abundance of straight lines and sharp edges in real-world scenes. Colonoscopy images (right) show centrally concentrated, more isotropic diffusion patterns, indicating a lack of directional geometric structures, with high-frequency energy decaying rapidly but influenced by specular highlights and unstructured noise.
   (b) Scatter plot of power-law spectral slope ($\alpha$) versus linearity of the fit ($R^2$).
   Natural street scenes (orange clusters) exhibit spectral slopes and linearity that strictly adhere to the characteristic power-law distribution of natural scenes. 
   Conversely, colonoscopy images (blue clusters) significantly deviate from this norm, demonstrating a distinct statistical shift in the frequency domain.
   This shift is characterized by impulsive high-frequency components derived from mucosal artifacts, which elevate the energy in the spectral tail and result in a slower decay rate. 
   The consistently lower $R^2$ values further indicate that biological tissue lacks the strict scale invariance typical of macroscopic natural scenes.
   }
\label{fig:intro:frequency}
\end{figure*}

We reassess the prevailing semantic hypothesis by identifying that the critical bottleneck lies in the low-level signal distribution rather than high-level abstraction.
This new assumption is substantively proved by our spectral domain analysis (see Fig. \ref{fig:intro:frequency}), which exposes a fundamental disconnection between the inductive bias of foundation models and the colonoscopic reality.
We observe that there exist a severe spectral collapse in colonoscopy: unlike the anisotropic harmonics of natural scenes, mucosal textures exhibit an isotropic energy void that fails to encode directional edge cues. 
Specifically, frequency bands crucial for inferring geometry, the structural signal energy in colonoscopy images is less than half of that found in natural scenes. This deficit is a deterministic physical artifact of colonoscopy, where tissue properties and optics inherently suppress high-frequency content. The result is a critical spectral bias when transferring foundation models to colonoscopy: the feature extractors of foundation models, tuned to a rich high-frequency signal, are presented with an input that falls below their effective activation threshold for depth reasoning. Consequently, the geometric failure of foundation models stems from a spectral incompatibility in the input signal, not from a semantic misunderstanding.

To solve this, we present a novel frequency-rectified adaptation paradigm, namely SpecDepth, to systematically reactivate these latent geometric capabilities.
The core of our framework is an adaptive wavelet modulation mechanism that functions as a spectral counter-force, exploiting the directional frequency selectivity of wavelet sub-bands to explicitly amplify the attenuated high-frequency signals essential for boundary delineation.
Crucially, to anchor this amplification in robust visual priors, we employ a hybrid encoder that freezes shallow layers while permitting deep-layer plasticity for frequency realignment.
Furthermore, to prevent the spectral enhancement from inadvertently magnifying high-frequency artifacts, we constrain the system with a geometry-aware regularization stream.
Ultimately, this synergistic design empirically validates our central thesis that transferability is hindered not by semantic misalignment but by spectral starvation, and that restoring low-level spectral consistency is a necessary precondition for unlocking the dormant geometric potential of foundation models.
The key contributions of this study can be summarized as follows: 

\begin{enumerate}[label=\roman*.]
  \item We identify the intrinsic spectral decay of colonoscopy as the fundamental barrier to transferability. 
  This finding shifts the adaptation paradigm from semantic retraining to statistical signal re-calibration.
  \item We propose SpecDepth, a framework driven by adaptive wavelet modulation. 
  This mechanism explicitly counteracts high-frequency attenuation to restore the structural cues necessary for precise boundary delineation.
  \item We introduce a multi-constraint objective combining scale-invariant regression and gradient smoothness. 
  This stabilizes the spectral amplification process to enhance anatomical coherence without magnifying sensor noise.
  \item We conduct extensive experiments on two public datasets, C3VD and SimCol3D, demonstrating state-of-the-art performance with minimal parameter overhead.
\end{enumerate}

\section{Related Work}

\subsection{Monocular Depth Estimation}

Monocular depth estimation has evolved from supervised learning to self-supervision and more recently to large-scale pre-training.
Early work started with multi-scale supervised regression, where \citet{Eigen2014} verified its feasibility on NYU Depth Dataset V2 \citep{Silberman2012} and KITTI dataset \citep{Geiger2012} and established a supervised baseline.
Subsequent work then improved accuracy with deeper convolutional architectures and attention-based representations \citep{conf/3DV/Laina2016, journal/IJMLC/Chen2021}.
Afterward, a self-supervised geometry-driven paradigm emerged, with SfMLearner jointly optimizing depth and pose using view reprojection as the learning signal \citep{conf/CVPR/Zhou2017}. 
Monodepth2 \citep{conf/ICCV/Godard2019} introduced minimum-reprojection and occlusion handling to improve robustness \citep{Godard2017b}.
These methods collectively established self-supervision as the mainstream choice when dense ground truth was unavailable.
Building on this trajectory, recent work used synthetic data and pseudo-labels for large-scale pretraining, exemplified by Depth Anything V2 with strong cross-domain generalization \citep{conf/NeurIPS/Yang2024}.
However, the performance of these methods degrades in colonoscopy due to low texture, specular highlights, and short working distances, which break photometric and textural assumptions.
Colonoscopy monocular depth estimation builds on the general pipeline but needs to tackles above mentioned domain-specific challenges.
For wide-angle colonoscopy, \citet{journal/FrontRobAI/Mathew2023} adapted the joint depth–pose framework to colonoscopy video and stabilized training with temporal modeling and occlusion handling.
To address boundary erosion and scale drift under low texture and variable lighting, geometry-aware approaches added geometric and scale constraints and enforced temporal consistency for self-supervised training \citep{conf/BIBM/Zhou2023,conf/MICCAI/Budd2024}.
Furthermore, to bridge the synthetic-to-real gap,  \citet{conf/MICCAI/Wang2024} introduced unsupervised adversarial domain adaptation, providing a way to reduce reliance on dense real annotations.
On this basis, EndoDAC and related methods migrated strong pre-trained depth backbones to surgical and colonoscopy scenes, using adapters or small parameter updates to improve generalization with few samples and across devices \citep{conf/MICCAI/Cui2024}.
For benchmarking and reproducibility, C3VD provides real paired depth from 2D to 3D registration \citep{journal/MedIA/Bobrow2023}, while SimCol3D \citep{journal/MedIA/Rau2024} offers a large-scale synthetic benchmark with official splits for systematic comparison.
Despite progress in temporal and geometric consistency, depth estimation of colonoscopy remains challenging due to its texture-sparse, reflective properties.

\subsection{Multi-Scale Frequency Domain Integration}

Multi-scale analysis plays a pivotal role in colonoscopy imaging by employing hierarchical decomposition to capture features at multiple scales, which addresses challenges like low resolution, motion artifacts, and uneven illumination to improve accuracy \citep{journal/IEEE/Unser1996,journal/SR/Yuan2024,journal/BSPC/Wang2025}.
Early multi-scale approaches, starting with spatial-domain methods like Gaussian or Laplacian pyramids, were widely used for image detail enhancement \citep{journal/EIVP/Jin2015,journal/RS/Tu2019}.
However, these spatial methods often struggle with precise frequency decoupling, making it difficult to separate structural textures from inherent noise.
Furthermore, more sophisticated tools like contourlet transforms have been adopted to fuse texture features across scales, indicating that frequency-domain enhancements are needed to better isolate signals from noise \citep{journal/TNNLS/Liu2021}.
Among various frequency-domain techniques, wavelet transforms have gained widespread prominence due to their capability to balance spatial details with noise suppression across different resolutions \citep{journal/PR/Tian2022,journal/SP/Yuan2025}.

Wavelet transform convolution has evolved from fixed multiscale decomposition to trainable frequency routing in CNNs for computer vision tasks \citep{journal/CIS/Zhao2023}.
Pioneered methods replaced pooling with the Discrete Wavelet Transform (DWT) and reconstructed with the inverse transform, which kept boundaries intact and reduced noise while limiting cost \citep{journal/Access/Liu2019}. 
In reconstruction, multi-level wavelet CNNs exploited subband priors to regularize sparse-view CT and improve fidelity \citep{journal/pm/Lee2020}.
After that, differentiable and learnable wavelet layers matured inside medical CNNs, extending subband processing into end to end training and validating subband priors in segmentation \citep{conf/BMVC/Hsieh2021}. 
For depth and disparity, Wavelet-MonoDepth predicted sparse wavelet coefficients in the decoder and reconstructed depth with the inverse transform, sharpening edges and improving efficiency \citep{conf/CVPR/Ramamonjisoa2021}. 
Then, Nested DWT CNN moved the transform into the encoder with learnable subbands to recover high-frequency content that downsampling discards \citep{journal/Sensors/Paul2023}.
Follow-up studies refine this trend by learning subband filters and attention weights, which enhanced fine-structure preservation under noise and domain shift.
Wavelet attention networks began this shift by reweighting subbands, so boundary evidence was emphasized in medical segmentation \citep{journal/BOE/Wang2022}. 
To move from weighting to learning, learnable wavelet pooling and differentiable DWT toolkits made the transform itself trainable and invertible inside CNNs \citep{conf/MICCAI/Cheng2024}.
Beyond attention, newer wavelet-guided segmentation models further refined subband usage to stabilize fine details without amplifying noise \citep{conf/IJCAI/Lu2025}.
Taken together, wavelet convolution has advanced from trainable operators to adaptive subband learning, proven effective across diverse medical imaging tasks \citep{conf/MLAI/Sriwichai2022,journal/MP/Qian2024,journal/Frontiers/Zhang2025} and showing strong potential for endoscopic applications \citep{journal/PO/Tan2024,conf/MICCAI/He2025}.
Most existing designs treat wavelet decomposition merely as a multi-scale representation with generic subband fusion and lack an explicit, learnable rectification step to selectively boost attenuated high-frequency subbands. 
Consequently, when the target domain is inherently deficient in strong textures and edges, these high-frequency cues remain weak, making the recovery of sharp structural boundaries difficult.

\subsection{Parameter-Efficient Adaptation of Foundation Models}

The adaptation of vision foundation models to medical imaging involves a fundamental trade-off between the retention of generalized visual priors and the acquisition of domain-specific semantics.
Conventional full fine tuning provides maximum plasticity but often results in the catastrophic forgetting of robust low-level features learned from massive datasets. 
This loss is particularly detrimental when the pre-trained weights capture universal geometric primitives that remain valid across domains regardless of semantic shifts.
To mitigate this risk, Parameter-Efficient Fine-Tuning methods like low-rank adaptation (LoRA) have been introduced to preserve priors by freezing the backbone and injecting a small number of new parameters~\citep{conf/ICLR/Hu2022,Biderman2024}.  
In medical vision, LoRA has demonstrated strong performance with minimal overhead~\citep{journal/CBM/Zhang2024,journal/IJCARS/Cui2024,conf/MICCAI/Cui2024,conf/ECCVW/SheikhZeinoddin2025}.
However the restrictive nature of these low-rank methods may limit the capacity of the model to learn complex high-rank features required for specific new tasks.
An alternative line of research explores hybrid or selective fine-tuning~\citep{gu2024build}. 
These strategies focus on identifying an optimal configuration that retains foundational knowledge in shallow layers while allowing deeper layers to specialize.
Crucially prior research suggests that shallow layers largely encode universal visual primitives such as frequency patterns and textures while deeper layers capture task-specific abstractions ~\citep{conf/NIPS/Raghu2019}.
This theoretical insight suggests that a selective freezing schedule could be particularly effective for domains with distinct low-level signal characteristics.

\begin{figure*}[t]\center
 \includegraphics[width=1\textwidth]{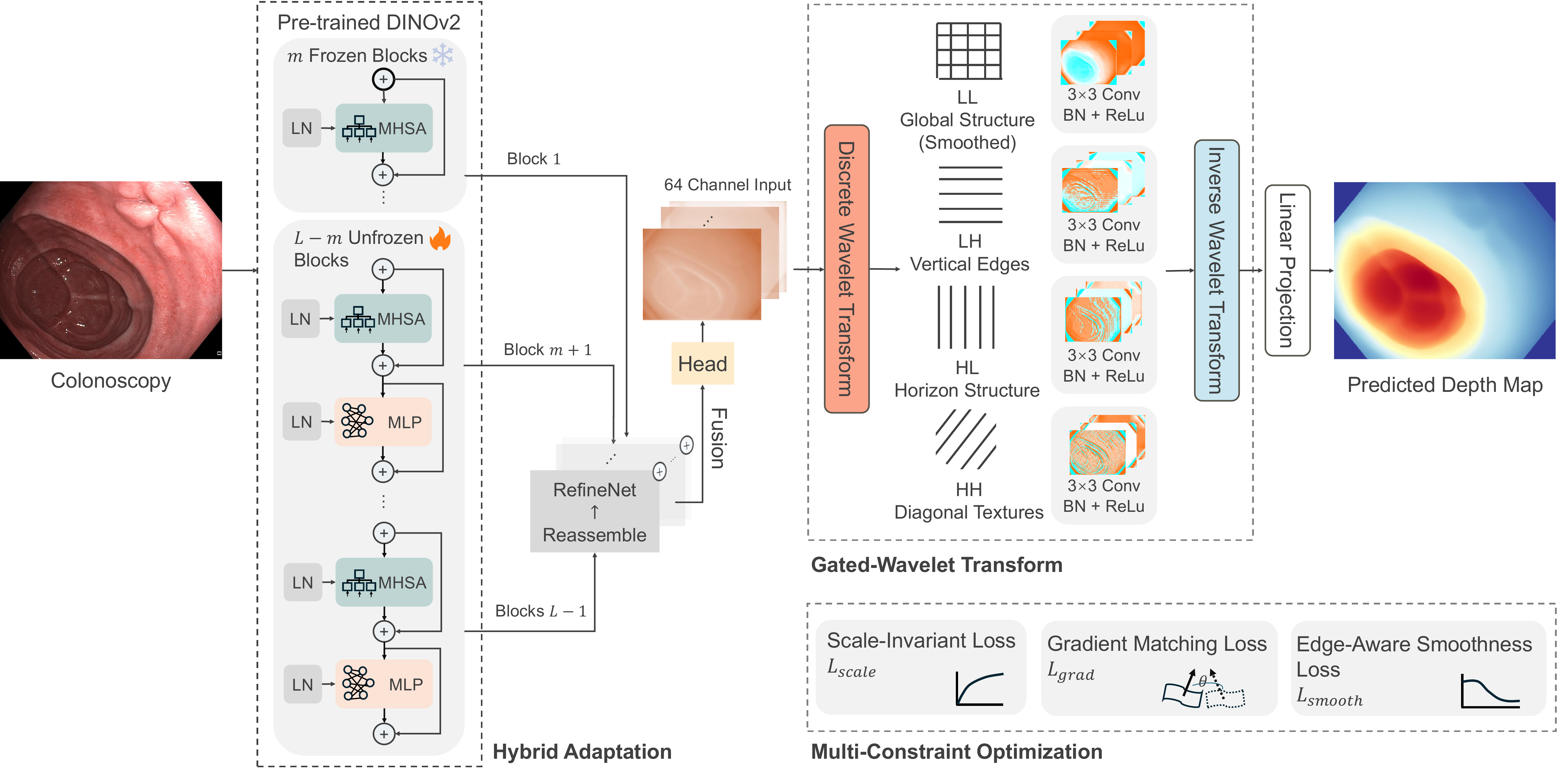}\\[-2ex]
   \caption{Overview of the proposed SpecDepth framework for the depth estimation of colonoscopy images. It employs a partially frozen DINOv2 encoder for stable feature extraction, followed by the gated wavelet  transform to decouple feature representations in the frequency domain and selectively amplify critical edge signals. LN: layer normalization. MHSA: multi-head self-attention. MLP: multi-layer perceptron. BN: batch normalization. }
\label{fig:method:framework}
\end{figure*}

\section{Methodology}

As illustrated in Fig. \ref{fig:method:framework}, our framework addresses the fundamental spectral mismatch encountered when transferring natural image foundation models to colonoscopy scenes. 
The core design divides the task into two distinct phases. 
The encoder focuses on preserving robust semantic representations while the decoder is responsible for rectifying the high frequency signal decay. 
We integrate a prior preserving adaptation strategy with a frequency rectified decoding mechanism to achieve accurate depth estimation in texture sparse environments.


\subsection{Hybrid Adaptation of Foundation Models with Preserved Shallow Priors}

The encoder of SpecDepth is built upon DINOv2 ViT-Small \citep{oquab2023dinov2}, a self-supervised vision transformer pre-trained on natural images. 
Given an input colonoscopy image $\mathbf{I} \in \mathbb{R}^{H \times W \times 3}$, it is first partitioned into a grid of $K$ non-overlapping patches. 
Each patch is linearly projected into a $D$-dimensional embedding and combined with positional encodings to form the input sequence $\mathbf{z}_0 \in \mathbb{R}^{K \times D}$. 
This sequence then propagates through $L$ consecutive Transformer blocks. 
Within each block $l$, the input $\mathbf{z}_{l-1}$ first passes through a layer normalization, followed by a multi-head self-attention (MHSA) mechanism. 
The output of MHSA is added back to the input via a residual connection to produce $\mathbf{z}'_l$. 
This intermediate representation is then normalized again and fed into a multi-layer perceptron, with another residual connection yielding the final output $\mathbf{z}_l$ of the block. 
Through this hierarchy, early layers encode low-level visual primitives such as edges, textures, and frequency patterns, while deeper layers capture increasingly semantic and task-specific concepts.

Adapting this foundation model to colonoscopy requires balancing the preservation of universal geometric priors against the need to learn domain-specific statistics. 
Full fine-tuning offers maximum plasticity but risks catastrophic forgetting of early-layer priors that remain valid across domains. 
Conversely, parameter-efficient methods such as LoRA assume task adaptations are inherently low-rank, which may be too restrictive for learning the complex, high-rank features required to interpret texture-sparse and deformable endoscopic anatomy. 
We therefore adopt a hybrid strategy: we freeze the first $m$ Transformer blocks and fully fine-tune the remaining $L-m$ blocks. 
The frozen shallow blocks serve as a stable ``scaffold'' that preserves fundamental visual structures (edges, contours, and frequency gradients, etc.), which are essential for geometric reasoning. 
The deeper blocks, granted full plasticity, adapt the representations to colonoscopy-specific statistics such as the appearance of mucosal folds, textureless regions, and specular artifacts. 
This configuration ensures that the encoder outputs $\mathbf{z}_L$ retain the robustness of general visual priors while incorporating domain-specific adaptations, providing a reliable foundation for the subsequent depth prediction.

\begin{figure*}[t]\center
  \includegraphics[width=0.8\textwidth]{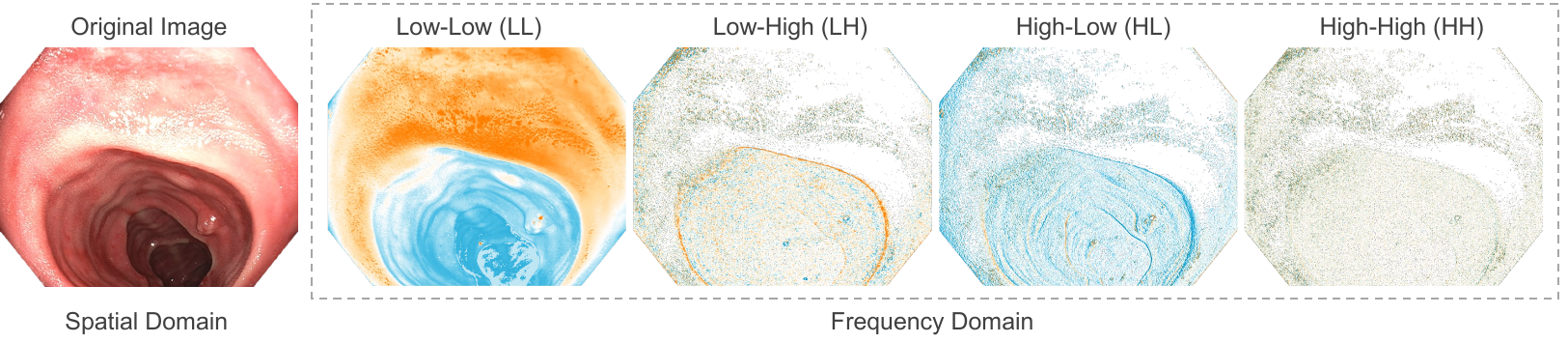}\\[-2ex]
    \caption{Illustration of wavelet decomposition process used in wavelet transform convolution. Note that here 2× upsampling applied to frequency-domain components for visual alignment with spatial domain. }
 \label{fig:method:wavelet}
 \end{figure*}

\subsection{Gated-Wavelet Spectral Rectification with Learnable Subband Weights} 

To obtain reliable geometric boundaries of colonoscopy, we rectify the frequency statistics of endoscopic features by rebalancing directional wavelet subbands.
Note that standard multi-scale convolutions can not explicitly disentangle these directional frequency components, which limits direct control over this spectral shift.
To explicitly modeling directional frequency subbands,  we introduce a gated-wavelet transform (GWT) in the dense prediction transformer head after multi-scale fusion, enabling boundary-aware spectral rectification without perturbing the pretrained encoder.
Specifically, given an input feature map $X \in \mathbb{R}^{C \times H \times W}$, we apply a two dimensional discrete wavelet transform to decompose it into four directional frequency subbands (see Fig. \ref{fig:method:wavelet}):
\begin{equation}
[X_{LL}, X_{LH}, X_{HL}, X_{HH}] = \mathrm{GWT}(X),
\label{eq2}
\end{equation}
where $X_{LL}$ retains low-frequency approximations encoding global structure, $X_{LH}$ and $X_{HL}$ highlight vertical and horizontal edges respectively, and $X_{HH}$ captures diagonal textures.
This decomposition offers directional boundary cues and frequency-specific control to correct collapsed and noisy high-frequency signals in colonoscopic features.

To implement a parameter-efficient rectification, we enhance each subband with a lightweight operator, consisting of a dedicated $3\times3$ convolution followed by batch normalization and ReLU activation, yielding $\widetilde{X}_{LL}$, $\widetilde{X}_{LH}$, $\widetilde{X}_{HL}$, and $\widetilde{X}_{HH}$.
Learnable global scalar weights $\boldsymbol{\omega} = [\omega_{LL}, \omega_{LH}, \omega_{HL}, \omega_{HH}]$ provide an explicit mechanism to rebalance subband contributions and to realign the decoder features with the inductive bias of the pretrained foundation model.
This choice intentionally keeps the modulation global and low dimensional, so the module performs frequency distribution realignment rather than pixel-level gating, which improves stability and reduces the risk of amplifying specular artifacts and sensor noise. 
The updated feature map is reconstructed through the inverse wavelet transform (IWT):
\begin{equation}
\hat{F} = \mathrm{IWT}\Big(
\begin{aligned}
&\omega_{LL} \cdot \widetilde{X}_{LL},\quad \omega_{LH} \cdot \widetilde{X}_{LH},\\
&\omega_{HL} \cdot \widetilde{X}_{HL},\quad \omega_{HH} \cdot \widetilde{X}_{HH}
\end{aligned}
\Big)
\label{eq3}
\end{equation}

This module realigns the overall frequency energy distribution of decoder features through global subband reweighting.
It compensates for attenuated high-frequency energy that suppresses geometric reasoning in colonoscopic scenes, while preventing artifact-driven high-frequency spikes from dominating the prediction. 
To ensure the rectification remains compatible with the pretrained priors, the rectified feature $\hat{F}$ is injected back into the decoder stream through residual addition with the original feature, and the fused representation is mapped to the depth prediction via the original prediction head.
This design is parameter efficient and does not require modifications to the pretrained backbone.

\subsection{Multi-Constraint Strategy for Structure-Aware Depth Estimation}

Spectral rectification strengthens high-frequency cues, but it can also risk amplifying noise that does not correspond to depth discontinuities.
Without additional constraints, the model can convert these non-geometric cues into incorrect depth discontinuities and noisy local depth variations.
We therefore propose a multi-constraint objective that complements GWT with explicit geometric supervision and edge-preserving regularization.
The objective preserves global scale, aligns depth discontinuities with geometry, and suppresses artifacts, achieved by three terms: $\mathcal{L}_\mathrm{scale}$ (scale-invariant structure), $\mathcal{L}_\mathrm{grad}$ (geometry-aligned discontinuities via gradient matching), and $\mathcal{L}_\mathrm{smooth}$ (edge-preserving smoothing).

\paragraph{Scale-Invariant Structure Loss}
Colonoscopic sequences often exhibit global scale variation due to camera motion and scene range.
To reduce sensitivity to this variation and emphasize relative structure, we adopt a scale-invariant logarithmic loss following Depth Anything v2~\citep{conf/NeurIPS/Yang2024}:
\begin{equation}
\begin{aligned}
\mathcal{L}_\mathrm{scale} =
& \sqrt{ \frac{1}{N} \sum_n (\log \hat{d}_n - \log d_n)^2 } - \lambda_\mathrm{s} \left( \frac{1}{N} \sum_n (\log \hat{d}_n - \log d_n) \right)^2
\end{aligned}
\label{eq4}
\end{equation}
where \(\hat{d}_n\) and \(d_n\) denote the predicted and ground-truth depth at pixel \(n\), \(N\) is the number of valid pixels, and $\lambda_\mathrm{s}$ is a balancing parameter.
This formulation penalizes scale-invariant structural errors while reducing sensitivity to global scale shifts.

\paragraph{Gradient Matching Loss}
In texture-sparse colonoscopic scenes, depth predictions often exhibit over-smoothed geometry and misaligned depth transitions even when the global structure is approximately correct.
While $\mathcal{L}_\mathrm{scale}$ constrains the overall structure up to scale, it does not explicitly enforce the spatial alignment and sharpness of depth transitions.
We therefore introduce a gradient matching loss that aligns first-order depth gradients between prediction and ground truth:
\begin{equation}
\begin{aligned}
\mathcal{L}_\mathrm{grad} =
& \frac{1}{N_x} \sum_{n \in \Omega_x} \left| \nabla_x \hat{d}_{n} - \nabla_x d_{n} \right| + \frac{1}{N_y} \sum_{n \in \Omega_y} \left| \nabla_y \hat{d}_{n} - \nabla_y d_{n} \right|
\end{aligned}
\label{eq5}
\end{equation}
where \( \Omega_x \) and \( \Omega_y \) denote the sets of valid pixel indices in the horizontal and vertical directions, respectively, and $\nabla$ denotes first-order finite differences computed at the output resolution.
This term reinforces the alignment of depth transitions with true geometry, promoting boundary-aware depth estimation even in regions with low intensity contrast.

\paragraph{Edge-Aware Smoothness Loss}
Although $\mathcal{L}_\mathrm{scale}$ and $\mathcal{L}_\mathrm{grad}$ provide direct supervision, they do not explicitly regularize the solution space in low-texture regions, where depth should remain smooth except at visually supported boundaries.
In these regions, emphasizing high-frequency cues can lead to unnecessary local variations in the predicted depth.
We therefore impose an edge-aware smoothness prior that regularizes depth in homogeneous areas while preserving discontinuities that are supported by the image evidence.
Let $I$ denote the input RGB image, and compute an intensity image $I^\mathrm{g}$ by converting $I$ to grayscale.
We then define per-pixel weights from image gradients so that smoothing is reduced near strong visual edges and strengthened in homogeneous regions:
\begin{equation}
\mathcal{L}_\mathrm{smooth} = \frac{1}{N} \sum_{n} \left( \alpha_x \cdot |\nabla_x \hat{d}_{n}| + \alpha_y \cdot |\nabla_y \hat{d}_{n}| \right),
\label{eq6}
\end{equation}
with dynamic weights:
\begin{equation}
\begin{aligned}
\alpha_x &= \exp\left( -\frac{1}{N} \sum_{n} \left| \nabla_x \hat{d}_{n} \right| \right),  
\alpha_y = \exp\left( -\frac{1}{N} \sum_{n} \left| \nabla_y \hat{d}_{n} \right| \right)
\end{aligned}
\label{eq8}
\end{equation}
The weights \( \alpha_x \) and \( \alpha_y \) are computed based on the mean gradient magnitude of the input image. 
This design preserves depth discontinuities at strong image edges, while suppressing small oscillations in texture-sparse regions where artifacts are most likely to be amplified.

\paragraph{Total Loss}
The total loss combines all these components:
\begin{equation}
\mathcal{L}_\mathrm{total} = \mathcal{L}_\mathrm{scale} + \lambda_\mathrm{grad} \cdot \mathcal{L}_\mathrm{grad} + \lambda_\mathrm{smooth} \cdot \mathcal{L}_\mathrm{smooth},
\label{eq9}
\end{equation}
where $\lambda_\mathrm{grad}$ and $\lambda_\mathrm{smooth}$ are balancing parameters.

\section{Experiments and Results}

\subsection{Data Acquisition and Pre-Processing}

For validation, we primarily evaluated SpecDepth on the public C3VD dataset~\citep{journal/MedIA/Bobrow2023}.
C3VD provides high-resolution colonoscopy video sequences with pixel-level ground-truth annotations including depth, surface normals, optical flow, occlusion masks, and camera trajectories. 
The dataset was collected using a clinical colonoscope and aligned through a precise 2D–3D registration pipeline, effectively overcoming the limitations of synthetic rendering in modeling realistic lighting and tissue appearance. 
Therefore, C3VD is particularly well-suited for evaluating depth estimation and 3D reconstruction methods in real-world colonoscopic settings.
Given its image quality and clinical realism, we prioritized C3VD in all experiments and used it exclusively for ablation studies.
Following the setup of prior work~\citep{Solano2025}, we divided C3VD into a 6344-frame training set, 1738-frame validation set, and 1268-frame test set, where the validation set was used specifically for conducting ablation experiments, ensuring comparability with previous benchmarks.

In addition, we conducted experiments on SimCol3D \citep{journal/MedIA/Rau2024} to assess the robustness and general applicability of our approach in another colonoscopic scenario. 
SimCol3D is a synthetic dataset from the MICCAI 2022 EndoVis challenge, containing over 36,000 colonoscopy images with corresponding depth annotations at a resolution of $475\times475$. 
We followed the official data split with a 28,776-frame training set and a 9,009-frame test set, consistent with the setup in MonoPCC~\citep{wang2025monopcc}. 
Unless otherwise specified, SpecDepth is trained and evaluated on SimCol3D under this in-domain setting (training on SimCol3D and testing on SimCol3D).

Note that we did not apply any dedicated masking algorithm for field-of-view black borders or specular highlights. 
Instead, invalid depth values are filtered using a simple range-based mask derived from the ground-truth depth. 
Specifically, pixels are considered valid only if the depth is positive and falls within a dataset-dependent interval $[d_{\min}, d_{\max}]$. 
This mask is applied consistently in both training (loss computation) and evaluation (metric computation), where out-of-range values are excluded rather than being clamped.

\subsection{Evaluation Metrics}

To comprehensively evaluate the performance of our depth estimation model, we adopt a set of standard monocular depth metrics commonly used in previous works~\citep{Solano2025, wang2025monopcc}. 
These metrics fall into two categories: 

1) Error-based metrics including AbsRel, SqRel, RMSE, RMSElog, log10, and SILog, which quantify the absolute or relative deviation between predicted and ground-truth depth.  

2) Accuracy-based metrics ($\delta_1$, $\delta_2$, $\delta_3$), which measure the percentage of pixels where the predicted depth falls within specified thresholds of the ground-truth depth.

The mathematical definitions of all employed metrics are summarized in Table~\ref{exp:tab:metrics}. 
We report all nine metrics when evaluating on the C3VD dataset, consistent with the evaluation protocol used in Col3D-MTL+SSL~\citep{Solano2025}. 
For the SimCol3D datasets, we follow the MonoPCC evaluation setting~\citep{wang2025monopcc}, and report only five error metrics: AbsRel, SqRel, RMSE, RMSElog, and $\delta_1$.

\begin{table}[t]
\footnotesize
\centering
\setlength{\tabcolsep}{1pt}
\caption{Depth estimation metrics used in this study.}
\label{exp:tab:metrics}
\begin{tabularx}{\linewidth}{@{}lX@{}}
\toprule
Metric & Formula \\
\midrule
AbsRel, SqRel & $\displaystyle \frac{1}{N} \sum_{n=1}^N \frac{|d_n - \hat{d}_n|}{d_n}$, $\displaystyle \frac{1}{N} \sum_{n=1}^N \frac{(d_n - \hat{d}_n)^2}{d_n}$ \\
RMSE, RMSElog & $\displaystyle \sqrt{\frac{1}{N} \sum_{n=1}^N (d_n - \hat{d}_n)^2}$, $\displaystyle \sqrt{\frac{1}{N} \sum_{n=1}^N (\log d_n - \log \hat{d}_n)^2}$\\
log10 & $\displaystyle \frac{1}{N} \sum_{n=1}^N \left| \log_{10} d_n - \log_{10} \hat{d}_n \right|$ \\
SILog & $\displaystyle \frac{1}{N} \sum_{n=1}^N \Delta_n^2 - \frac{1}{N^2} \left( \sum_{n=1}^N \Delta_n \right)^2$, where $\displaystyle \Delta_n = \log d_n - \log \hat{d}_n$ \\
$\delta_{1/2/3}$ & \% pixels s.t. $\displaystyle \max\left(\frac{d_n}{\hat{d}_n}, \frac{\hat{d}_n}{d_n} \right) < 1.25/1.25^2/1.25^3$ \\
\bottomrule
\end{tabularx}
\end{table}

\begin{table*}[t]
\footnotesize
  \caption{Summary of quantitative results of different method for depth estimation on the C3VD dataset. }
  \label{exp:tab:C3VD_compare}
  \centering
  \setlength{\tabcolsep}{3pt}
  \begin{tabular}{lcccccccccc} 
    \toprule
    Method & Params (M) & AbsRel↓ & SqRel↓ & log10↓ & RMSE↓ & RMSElog↓ & SILog↓ & $\delta_1 \uparrow$ & $\delta_2 \uparrow$ & $\delta_3 \uparrow$ \\
    \midrule
    NeWCRFs \citep{conf/CVPR/Yuan2022} & 270.4 & 0.133 & 0.557 & 0.048 & 3.058 & 0.151 & 11.817 & 0.854 & 0.973 & 0.997\\
    NDDepth \citep{journal/TPAMI/Shao2024} & 348.4 & 0.316 & 2.651 & 0.111 & 7.043 & 0.322 & 24.255 & 0.498 & 0.873 & 0.956 \\
    BTS \citep{lee2019big}  & 50.3 & 0.127 & 0.622 & 0.055 & 3.823 & 0.150 & 11.400 & 0.812 &0.979 &0.998 \\
    Col3D-MTL+SSL \citep{Solano2025} & 50.3 & 0.107 & 0.346 & 0.042 & 2.729 & 0.128 & 10.072 & 0.899 & 0.989 & \textbf{0.999}  \\
    \midrule
    EcoDepth \citep{conf/CVPR/Patni2024} & 993.48 & 0.103 & 0.367 & 0.042 & 2.721 & 0.116 & 8.539 & 0.901 & 0.994 & 0.998\\
    Marigold \citep{conf/CVPR/ke2023} & 865.92 & 0.075 & 0.228 & 0.032 & 2.353 & 0.115 & 11.424 & 0.933 & 0.988 & 0.995 \\
    \hdashline
    Depth Anything V2 \citep{Yang2024cvpr} & \textbf{24.8} & 0.104 & 0.519 & 0.044 & 6.312 & 0.146 & 0.132 & 0.894 & 0.997 & \textbf{0.999} \\
    \textbf{SpecDepth (Ours)} & 26.0 & \textbf{0.022} & \textbf{0.088} & \textbf{0.009} & \textbf{1.957} & \textbf{0.039} & \textbf{0.037} & \textbf{0.997} & \textbf{0.999} & \textbf{0.999} \\
    \bottomrule
  \end{tabular}
\end{table*}

\begin{table*}[t]
\footnotesize
  \caption{Summary of quantitative results of different method for depth estimation on the SimCol3D dataset. } 
  \label{exp:tab:SimCol3D_compare}
  \centering
  \setlength{\tabcolsep}{6pt}
  \begin{tabular}{lccccc} 
    \toprule
    Method & AbsRel↓ & SqRel↓ & RMSE↓ & RMSElog↓ & $\delta_1 \uparrow$ \\
    \midrule
    Monodepth2 \citep{conf/ICCV/Godard2019}       & 0.076 & 0.061 & 0.402 & 0.106 & 0.950 \\
    FeatDepth \citep{conf/ECCV/shu2020}       & 0.077 & 0.069 & 0.374 & 0.098 & 0.957 \\
    HR-Depth \citep{conf/AAAI/lyu2021}       & 0.072 & 0.044 & 0.378 & 0.100 & 0.961 \\
    DIFFNet \citep{conf/BMVC/zhou2021}         & 0.074 & 0.053 & 0.401 & 0.105 & 0.957 \\
    Endo-SfMLearner \citep{journal/MIA/ozyoruk2021}  & 0.072 & 0.042 & 0.407 & 0.103 & 0.950 \\
    AF-SfMLearner \citep{journal/MIA/shao2022}    & 0.071 & 0.045 & 0.372 & 0.099 & 0.961 \\
    MonoViT ~\citep{conf/3DV/zhao2022}          & 0.064 & 0.034 & 0.377 & 0.094 & 0.968 \\
    Lite-Mono \citep{conf/CVPR/Zhang2023LiteMono}        & 0.076 & 0.050 & 0.424 & 0.110 & 0.950 \\
    MonoPCC \citep{wang2025monopcc}          & 0.058 & 0.028 & 0.347 & 0.090 & 0.975 \\
    \midrule
    EcoDepth \citep{conf/CVPR/Patni2024}          & 0.084 & 0.048 & 0.484 & 0.120 & 0.923 \\
    Marigold \citep{conf/CVPR/ke2023}          & 0.043 & 0.098 & 1.190 & 0.082 & 0.970 \\
    \hdashline
    Depth Anything V2 \citep{Yang2024cvpr}  &0.037 &0.033 &0.451 &0.071 &0.979 \\
    \textbf{SpecDepth (Ours)} & \textbf{0.027} & \textbf{0.023} & \textbf{0.335} & \textbf{0.060} & \textbf{0.984} \\
    \bottomrule
  \end{tabular}
\end{table*}

\subsection{Implementation Details}

The framework was implemented in PyTorch using a single NVIDIA RTX 4090 GPU. 
The encoder was initialized with the ViT-Small variant of DINOV2~\citep{oquab2023dinov2}, a model with 26M parameters.
During training, the input resolution was set to $320 \times 320$ for the C3VD dataset, while a higher resolution of $475 \times 475$ was used for SimCol3D.
Hyperparameters were set as below: $m=2$, $\lambda_\mathrm{s}=0.5$, $\lambda_\mathrm{grad}$ = 0.1, and $\lambda_\mathrm{smooth}$ = 0.1. 
The model was trained for approximately 10 epochs on C3VD and 30 epochs on SimCol3D with a batch size of 16.
Optimization was performed using AdamW with an initial learning rate of $5\times10^{-6}$ and 5000 warm-up steps.
Unless otherwise specified, we used the same optimizer settings, learning rate schedule, and training epochs for all ablation experiments.
In addition to our proposed method, we included a baseline that directly fine-tuned Depth Anything V2 (ViT-Small) under the same training and evaluation protocol, ensuring a controlled comparison.
For other comparison methods, we reported the results from their original papers under the corresponding protocols. 
On the C3VD dataset, we strictly followed the training and test split as well as the evaluation protocol from \citet{Solano2025}. 
On the SimCol3D dataset, we aligned with the experimental settings and metric selection of \citet{wang2025monopcc}. 
This allows us to directly reuse the reported results and ensure consistent experimental conditions across all comparisons.

\begin{figure*}[t]\center
\includegraphics[width=1\textwidth]{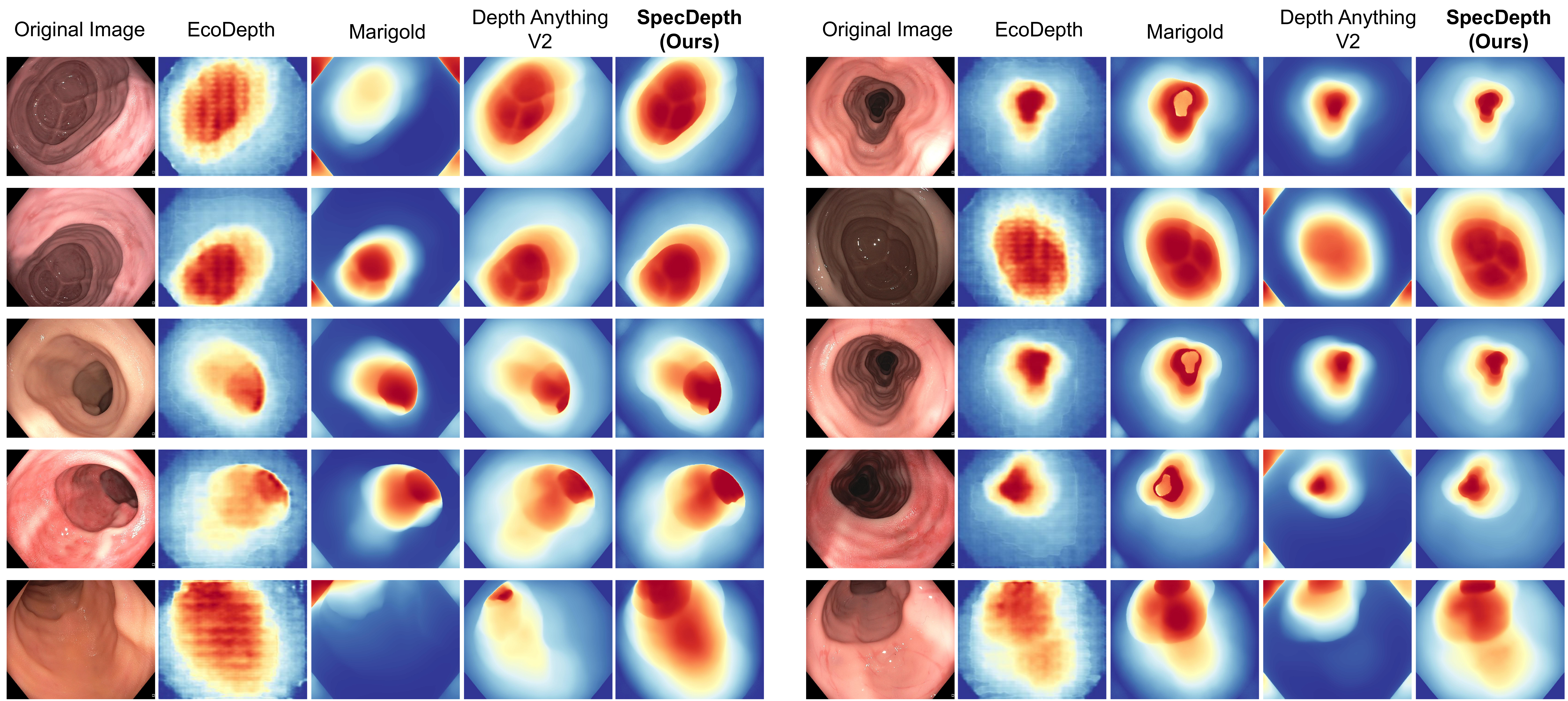}\\[-2ex]
   \caption{Visualization of monocular depth estimation on the C3VD dataset of different methods. C3VD is a real clinical colonoscopy dataset providing high-resolution video sequences with pixel-level ground-truth depth annotations. Compared to the baseline methods, SpecDepth demonstrated superior preservation of structural boundaries and produces smoother, more geometrically coherent gradients in texture-sparse mucosal regions.
   } 
\label{exp:result:visual_C3VD}
\end{figure*}

\begin{figure*}[t]\center
\includegraphics[width=0.98\textwidth]{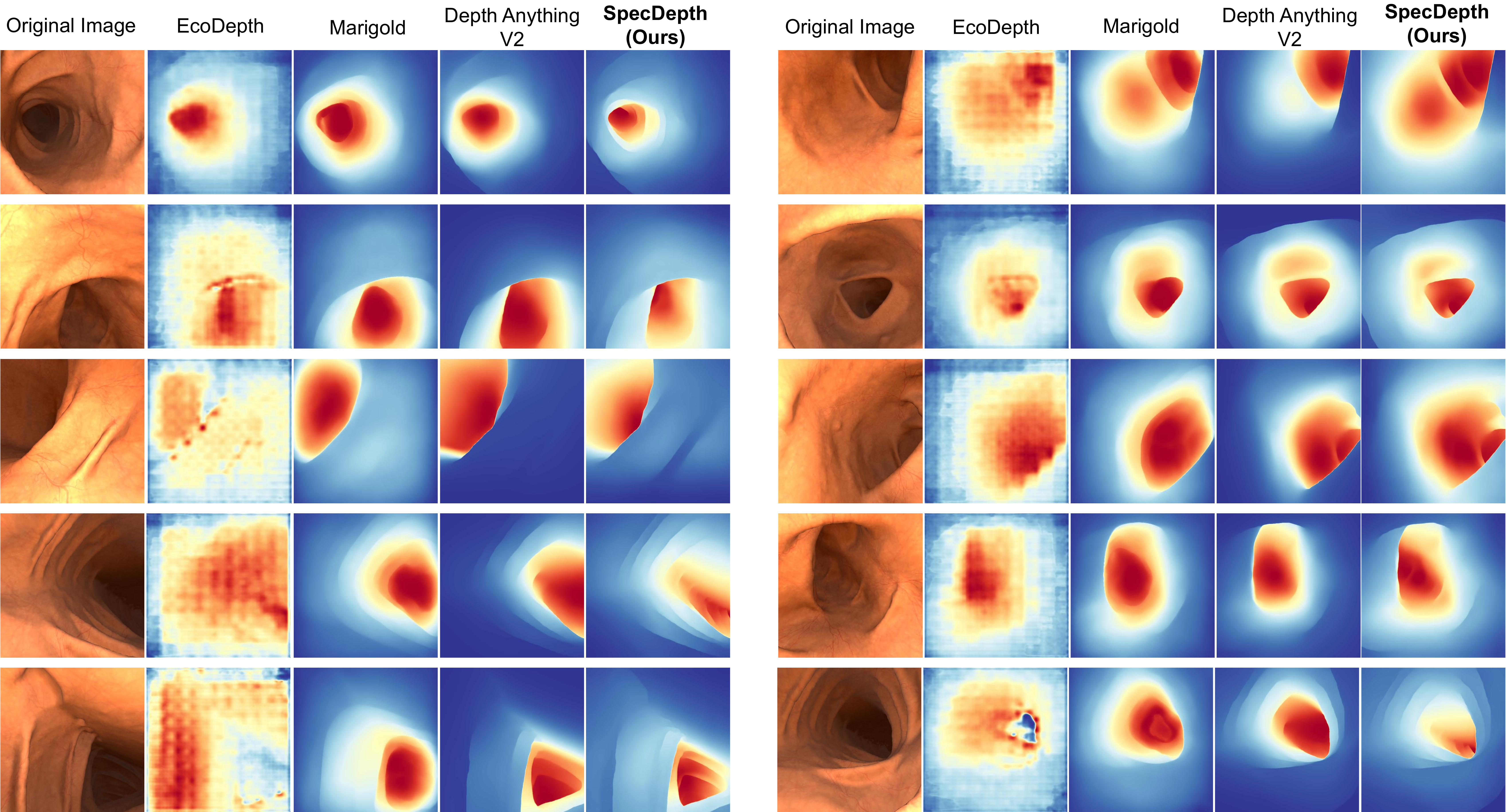}\\[-2ex]
   \caption{Visualization of monocular depth estimation on the SimCol3D dataset of different methods. 
   EcoDepth was heavily corrupted by structured noise, while Marigold, though smoother, tended to produce blob-shaped depth maps that failed to follow the elongated recession of the lumen. Depth Anything V2 captured the overall scene geometry more faithfully but often folded boundaries and transitional regions. SpecDepth produced the most coherent results, accurately rendering the tubular depth gradient along the lumen axis and maintaining crisp fold structures throughout.} 
\label{exp:result:visual_Simcol}
\end{figure*}

\subsection{Comparison Study}

We evaluate SpecDepth against three categories of methods on two public colonoscopy datasets (C3VD and SimCol3D; Tables~\ref{exp:tab:C3VD_compare} and \ref{exp:tab:SimCol3D_compare}): 
(i) the general-purpose foundation model (Depth Anything V2 \citep{Yang2024cvpr}), which reveal the performance gap when directly applying natural-image pre-trained models to colonoscopic domains; 
(ii) task-specific depth estimators (EcoDepth \citep{conf/CVPR/Patni2024} and Marigold \citep{conf/CVPR/ke2023}), which represent state-of-the-art designs for general-domain depth but lack colonoscopic adaptation; and 
(iii) prior colonoscopy-specific methods (e.g., Col3D-MTL+SSL \citep{Solano2025}, MonoPCC \citep{wang2025monopcc}), which establish the existing benchmark for this task.
SpecDepth achieved the best overall performance on both datasets, with particularly evident gains on C3VD: an AbsRel of 0.022 and $\delta_1$ of 0.997, significantly improving upon both the foundation model baseline (Depth Anything V2: 0.104 AbsRel, 0.894 $\delta_1$) and the best task-specific estimator (Marigold: 0.075 AbsRel, 0.933 $\delta_1$). 
The concurrent reduction in RMSE (1.957 vs. 2.353) and RMSElog (0.039 vs. 0.115) confirmed that SpecDepth eliminated large outliers while preserving scale consistency, a critical requirement for texture-sparse mucosal surfaces where baselines over-smooth folds and blur depth transitions. 
Notably, the gap between Depth Anything V2 and SpecDepth underscored that direct fine-tuning of foundation models, without targeted frequency adaptation, remained insufficient for colonoscopic domains. 
Conversely, the improvement over Marigold and EcoDepth demonstrated that even sophisticated general-domain depth architectures cannot bridge the spectral domain gap without explicit high-frequency rectification.
On SimCol3D, SpecDepth maintained its advantage (0.027 AbsRel, 0.984 $\delta_1$), surpassing both task-specific estimators (Marigold: 0.043 AbsRel; EcoDepth: 0.084 AbsRel) and prior colonoscopy-specific methods (MonoPCC: 0.058 AbsRel). 
This cross-dataset consistency, despite the different image statistics and synthetic origin of SimCol3D, validated that spectral rectification addressed a fundamental property of colonoscopic imagery rather than exploiting dataset-specific artifacts. 
The fact that Depth Anything V2 (natural images) and EcoDepth/ Marigold (general-depth) both underperformed relative to SpecDepth reinforces our core claim: 
the primary barrier is not semantic but spectral, and our frequency-domain intervention provides a transferable adaptation mechanism that generalizes across colonoscopic datasets.

Qualitative results in Figures \ref{exp:result:visual_C3VD} and \ref{exp:result:visual_Simcol} corroborate the quantitative findings. 
On the C3VD dataset (Fig. \ref{exp:result:visual_C3VD}), which captures real clinical colonoscopy scenarios with complex mucosal structures, EcoDepth exhibited significant depth artifacts, particularly in regions with specular reflections and along fold boundaries where it produced fragmented depth estimates. 
Marigold, while generating smoother outputs, tended to over-smooth critical anatomical features, blending together distinct folds and failing to preserve the sharp transitions that define colon wall topography. 
Depth Anything V2 demonstrated better global scene comprehension but struggled with depth discontinuities at fold boundaries, often bleeding depth values between foreground and background structures. 
In contrast, SpecDepth produced depth maps that maintained crisp edge delineation at fold boundaries while ensuring smooth, physically-plausible depth gradients in uniform mucosal regions. 
This is particularly evident in texture-sparse areas where other methods introduced noise or artificial gradients.
The SimCol3D results (Fig. \ref{exp:result:visual_Simcol}) further emphasize these differences, especially in handling the characteristic tubular geometry of the colon.
The predictions of EcoDepth were heavily corrupted by structured noise that followed specific patterns, suggesting an over-reliance on texture cues rather than geometric understanding. 
Marigold produced blob-shaped depth maps that failed to capture the elongated recession of the lumen, compressing the natural depth progression along the colon axis. Depth Anything V2 captured the overall scene geometry more faithfully but often folds depth boundaries in transitional regions, creating unnatural depth inversions at fold intersections. 
SpecDepth uniquely rendered the tubular depth gradient with accurate perspective scaling along the lumen axis while preserving the intricate structure of individual folds, capturing both the global scene geometry and local anatomical details.
Together, these results establish SpecDepth as a new state-of-the-art for monocular depth estimation in colonoscopy, with consistent gains across multiple challenging datasets.

\subsection{Ablation Study} \label{exp:ablation study}

\begin{table}[t]
  \caption{Ablation study of the proposed model on the C3VD validation set. GWT: gated-wavelet transform. MC: multi-constraint.}
  \label{exp:tab:ablation}
  \centering
  \footnotesize
  \resizebox{\columnwidth}{!}{
  \setlength{\tabcolsep}{3pt}
  \begin{tabular}{@{}lll cccc@{}} 
    \toprule
    Encoder & GWT & MC & AbsRel ↓ & SqRel ↓ & RMSE ↓ & $\delta_1 \uparrow$ \\
    \midrule
    Full Fine-Tuning & $\times$ & $\times$ & 0.125 & 0.571 & 5.205 & 0.866 \\
    Full Fine-Tuning & $\checkmark$ & $\times$ & 0.104 & 0.578 & 4.461 & 0.939 \\
    Full Fine-Tuning & $\times$ & $\checkmark$ & 0.092 & 0.562 & 5.517 & 0.881 \\
    Full Fine-Tuning & $\checkmark$ & $\checkmark$ & 0.077 & 0.553 & 3.703 & 0.960 \\
    LoRA & $\times$ & $\times$ & 0.133 & 0.688 & 5.862 & 0.859 \\
    LoRA & $\checkmark$ & $\times$ & 0.086 & 0.567 & 3.866 & 0.953 \\
    LoRA & $\checkmark$ & $\checkmark$ & 0.082 & 0.559 & 3.622 & 0.958 \\
    Hybrid Strategy & $\times$ & $\times$ & 0.128 & 0.707 & 6.551 & 0.839 \\
    Hybrid Strategy & $\checkmark$ & $\times$ & 0.086 & 0.550 & 3.642 & 0.965 \\
    Hybrid Strategy & $\times$ & $\checkmark$ & 0.114 & 0.558 & 5.211 & 0.874 \\
    Hybrid Strategy & $\checkmark$ & $\checkmark$ & \textbf{0.076} & \textbf{0.545} & \textbf{3.441} & \textbf{0.970} \\
    \bottomrule
  \end{tabular}
  }
\end{table}

\begin{table}[t]
  \caption{Sensitivity analysis of the gradient and smoothness loss weights on the C3VD validation set.}
  \label{exp:tab:hyperparameter}
  \centering
  \footnotesize
  \setlength{\tabcolsep}{2.5pt}
  \begin{tabular}{@{}ll cccc@{}}
    \toprule
    $\lambda_{\mathrm{grad}}$ &
    $\lambda_{\mathrm{smooth}}$ &
    AbsRel~$(\downarrow)$ &
    SqRel~$(\downarrow)$ &
    RMSE~$(\downarrow)$ &
    $\delta_1 \uparrow$ \\
    \midrule
    0   & 0   & 0.125 & 0.571 & 5.205 & 0.866 \\
    0   & 0.1 & 0.115 & 0.571 & 5.310 & 0.876 \\
    0   & 0.2 & 0.112 & 0.570 & 5.308 & \textbf{0.881} \\
    \midrule
    0.1 & 0   & 0.099 & 0.566 & 5.694 & 0.879 \\
    0.1 & 0.1 & \textbf{0.092} & \textbf{0.562} & 5.517 & \textbf{0.881} \\
    0.1 & 0.2 & 0.122 & 0.572 & 5.213 & 0.863 \\
    \midrule
    0.2 & 0   & 0.105 & 0.568 & 5.403 & 0.873 \\
    0.2 & 0.1 & 0.112 & 0.570 & 5.402 & 0.873 \\
    0.2 & 0.2 & 0.112 & 0.568 & 5.365 & 0.868 \\
    \bottomrule
  \end{tabular}
\end{table}

We evaluated SpecDepth to assess the individual and synergistic contributions of its two core components: the GWT for spectral rectification and the multi-constraint (MC) objective for geometric regularization. 
All ablations were conducted on the C3VD validation split. 
As shown in Table~\ref{exp:tab:ablation}, GWT served as the primary driver of transferable geometric priors: integrating it consistently improves performance across all adaptation strategies, including full fine-tuning, LoRA, and the hybrid strategy, demonstrating that the gains stemmed from rectified high-frequency cues rather than a specific parameter update scheme. 
Notably, the improvements were most pronounced under the hybrid strategy, where backbone updates were most constrained, suggesting that GWT addressed a fundamental bottleneck common to all fine-tuning approaches in the colonoscopy domain. 
The hybrid strategy itself proved to be an effective baseline by decoupling preserved backbone priors from adaptable low-level statistics; 
while its naive form underperformed unconstrained fine-tuning, it became the most competitive configuration when augmented with GWT and MC, indicating that targeted spectral intervention obviated the need for extensive parameter updates. 
The MC module further converted restored high-frequency structure into geometrically faithful depth by penalizing local inconsistencies. 
When applied alone, MC yielded measurable gains; when combined with GWT, it delivered additional improvement, confirming that spectral rectification alone cannot fully constrain local shape errors on challenging deformable and specular tissues. 
By enforcing alignment between depth discontinuities and image structure while maintaining smoothness elsewhere, MC mitigated the risk of amplifying non-geometric high-frequency artifacts. 
The full model thus achieved not only lower average error (AbsRel: 0.022) but also fewer large deviations, as reflected by concurrent gains in $\delta_1$ (0.997) and RMSE (1.957).
It revealed that spectral recovery must be paired with geometric regularization to translate enhanced cues into reliable depth predictions.

\begin{figure*}[t]\center
\includegraphics[width=1\textwidth]{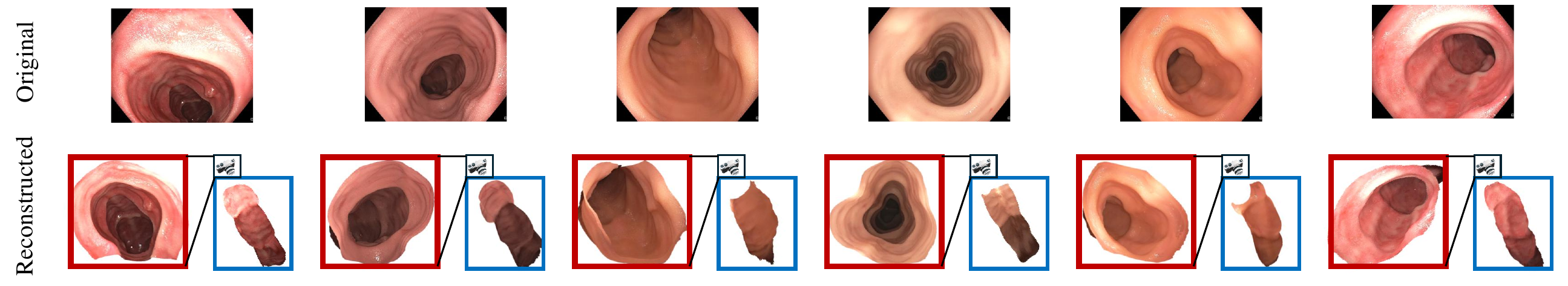}\\[-2ex] 
   \caption{Qualitative results of 3D point cloud reconstruction on the C3VD dataset. The top row displays the original colonoscopy image. The bottom row presents reconstructed point cloud generated by Open3D library.} 
\label{exp:result:3d_reconstruction}
\end{figure*}

We further analyzed the importance of the proposed loss and the sensitivity of the loss weighting parameters $\lambda_{\mathrm{grad}}$ and $\lambda_{\mathrm{smooth}}$ under the hybrid encoder setting with GWT-augmented decoder (see Table~\ref{exp:tab:hyperparameter}). 
The configuration $(\lambda_{\mathrm{grad}}, \lambda_{\mathrm{smooth}}) = (0.1, 0.1)$ achieved the best overall performance, yielding the lowest AbsRel (0.092) and SqRel (0.562) while maintaining competitive $\delta_1$ (0.881). 
This balance aligned with our design objective: improving global geometric accuracy without sacrificing edge fidelity. 
Notably, setting either weight to zero degrades performance, omitting $\mathcal{L}_\mathrm{grad}$ increased AbsRel by 25\% (0.115 vs. 0.092), while removing $\mathcal{L}_\mathrm{smooth}$ elevated SqRel, indicating that both terms were essential for suppressing noise and anchoring depth discontinuities to true geometry. 
Excessively high weights (0.2) also proved detrimental, suggesting that over-regularization can suppress legitimate high-frequency structure alongside artifacts. 
Note that our complementary analysis of the learned wavelet gates revealed that training increased the high-frequency sub-band activations ($g_{HH}{:}g_{HL}{:}g_{LH} = 1.05{:}1.04{:}1.01$) while slightly decreasing the low-frequency gate ($g_{LL}=0.98$), relative to uniform initialization. 
This shift directly linked the quantitative gains in Table~\ref{exp:tab:ablation} to the intended spectral rectification behavior, confirming that SpecDepth improved depth estimation by restoring the high-frequency cues necessary for engaging the pretrained geometric priors.

\begin{figure*}[t]\center
\includegraphics[width=1\textwidth]{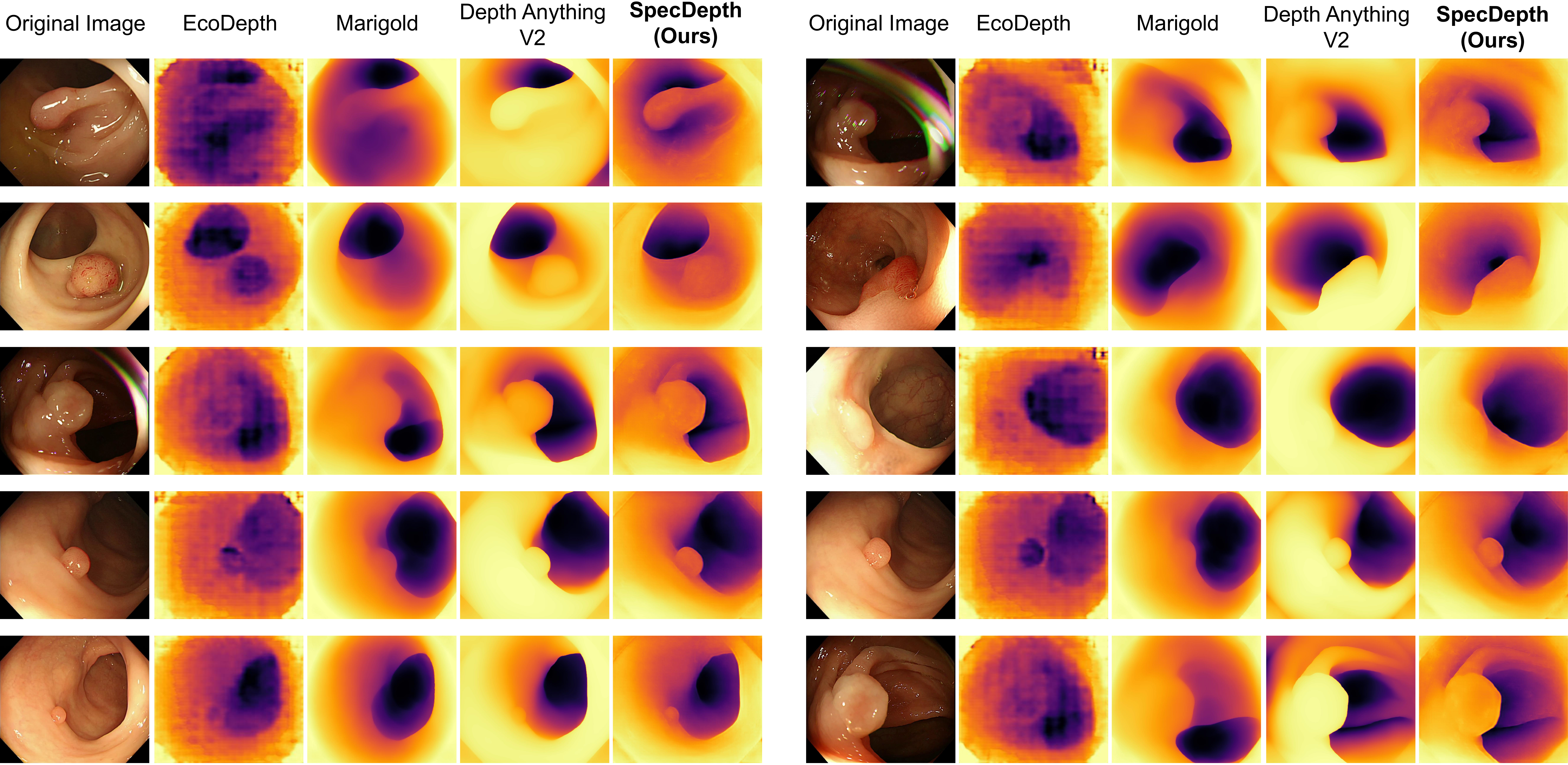}\\[-2ex]
   \caption{Visualization of monocular depth estimation on colorectal polyp scenes of different methods, evaluated in a zero-shot setting without any fine-tuning. 
   Images are frames extracted from the Polyp-Size dataset, a high-resolution colonoscopy video collection featuring colorectal polyps of diverse morphologies. 
   While EcoDepth struggled with visible grid-like artifacts and Marigold tended to dissolve polyp boundaries into the surrounding mucosa, Depth Anything V2 recovered a more coherent global layout yet still lost the three-dimensional relief of lesions. 
   SpecDepth preserved the convex profile of each polyp with well-defined boundary transitions and smoothly varying surface gradients, demonstrating more robust generalization across diverse polyp morphologies.} 
\label{exp:result:visual_polyp}
\end{figure*}

\begin{figure*}[t]\center
\includegraphics[width=1\textwidth]{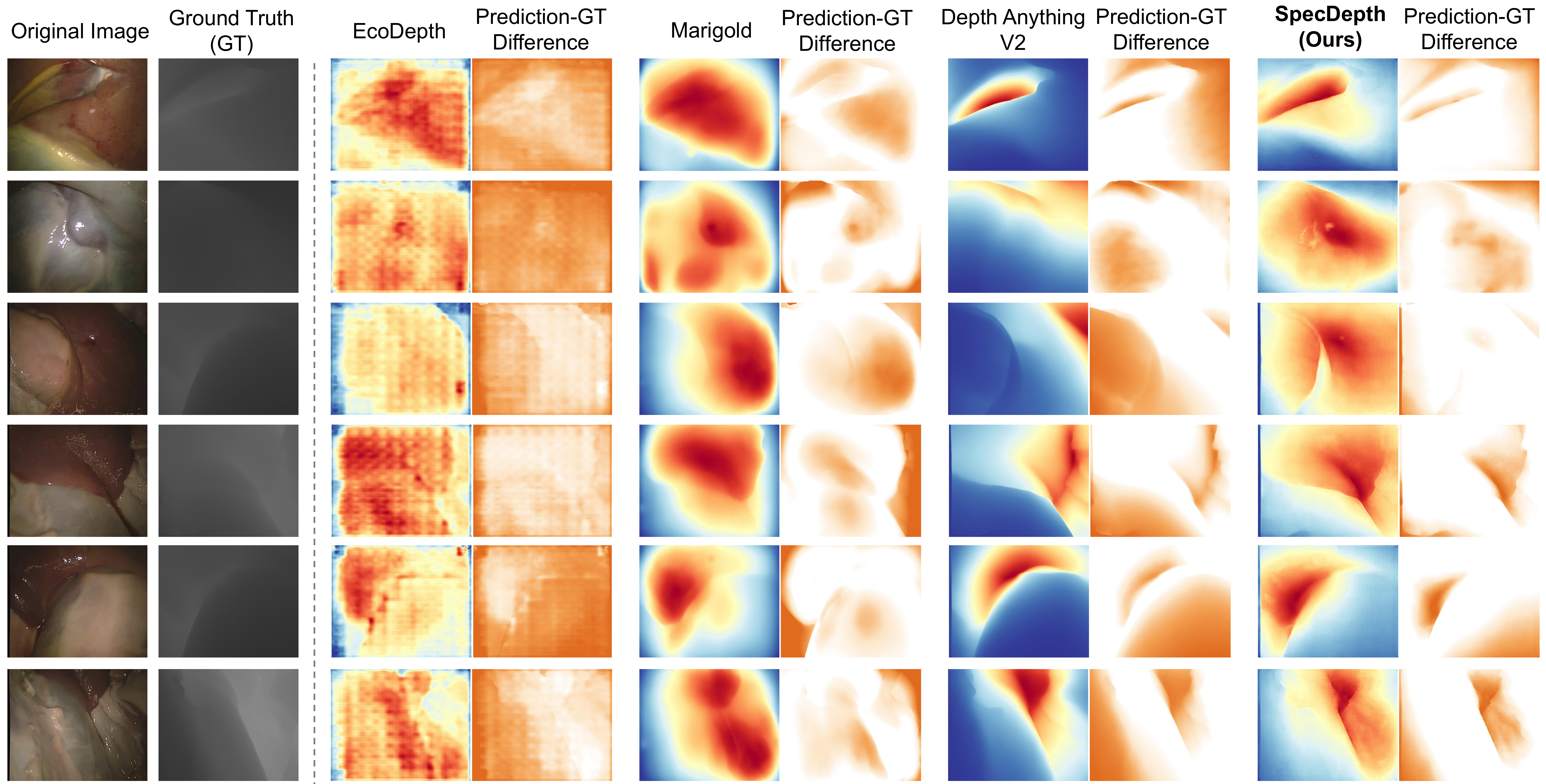}\\[-2ex]
   \caption{\textbf{Visualization of monocular depth estimation on the SERV-CT dataset of different methods, evaluated in a zero-shot setting without any fine-tuning.}
   SERV-CT is a laparoscopic stereo-endoscopic dataset capturing ex vivo porcine abdominal tissue. 
   Ground-truth depth maps are derived by aligning cone-beam CT scans with the corresponding stereo endoscope images through constrained manual registration.
   Each predicted depth map is paired with an absolute "Prediction-GT" difference map, where deep amber regions highlight large errors, and white areas indicate minimal error.
   SpecDepth produced difference maps with significantly more white areas compared to the other baselines, demonstrating superior zero-shot generalization and accuracy in complex endoscopic scenes.} 
\label{exp:result:visual_SERV_CT}
\end{figure*}

\subsection{Monocular 3D Reconstruction from Estimated Depth}

To demonstrate the practical utility of SpecDepth beyond static depth estimation, we evaluated its application to monocular 3D reconstruction in colonoscopy, a challenging task due to low texture, deformable tissues, and dynamic scenes. 
As shown in Fig.~\ref{exp:result:3d_reconstruction}, by leveraging its predicted depth maps and corresponding camera intrinsic and extrinsic parameters, SpecDepth directly generates high-fidelity dense surface models.
In smooth mucosal regions, the reconstructions were continuous and artifact-free; around complex structures such as intestinal folds, they preserved fine curvature and sharp boundaries, which are essential properties for maintaining anatomical plausibility. 
Reconstructions from multiple viewpoints exhibited consistent alignment with the original image geometry, confirming that the recovered depth encoded coherent 3D structure rather than view-specific artifacts.
In contrast to baseline methods that tended to over-smooth or distort depth in texture-sparse regions, SpecDepth delivered reliable geometric accuracy without additional fine-tuning or auxiliary inputs. 
These results suggested that spectral rectification not only improved depth metrics but also translated to tangible gains in 3D scene understanding, positioning SpecDepth as a foundational component for real-time colonoscopic navigation and downstream clinical applications.

\subsection{External Validation}

To rigorously evaluate the generalization capability of our proposed model beyond its training distribution, we conducted zero-shot testing on two distinctly different endoscopic datasets: (1) colorectal polyp scenes from the Polyp-Size dataset \citep{journal/SD/song2025polyp}, a high-resolution colonoscopy video collection featuring polyps of diverse morphologies, and (2) laparoscopic surgery scenes from the SERV-CT dataset \citep{journal/MedIA/edwards2022}, which captures abdominal cavity views with different instrumentation and tissue characteristics. 
The Polyp-Size dataset lacks ground-truth depth annotations, and SERV-CT, though annotated, is too small in scale for statistically meaningful quantitative evaluation.
Therefore, we rely on qualitative visual analysis to assess cross-domain transferability, a challenging test of whether the model has learned genuine geometric priors rather than dataset-specific shortcuts.

Figure \ref{exp:result:visual_polyp} presents zero-shot depth predictions on colorectal polyp scenes, where each method reveals distinct generalization behaviors when encountering unseen polyp morphologies. 
EcoDepth exhibited pronounced grid-like artifacts that corrupt the polyp surface, suggesting overfitting to texture patterns in its training data rather than learning robust shape priors. 
Marigold produced smoother outputs but consistently dissolved polyp boundaries into the surrounding mucosa, failing to delineate the critical transition between lesion and healthy tissue, a fundamental requirement for clinical utility. 
Depth Anything V2 recovered a more coherent global layout of the scene but still lost the three-dimensional relief of polyps, flattening their convex profiles into subtle undulations. 
In striking contrast, SpecDepth preserved the characteristic convex protrusion of each polyp with well-defined boundary transitions and smoothly varying surface gradients, accurately capturing both the overall protrusion and the fine surface texture that clinicians use to assess lesion morphology.

The SERV-CT laparoscopic results (Fig. \ref{exp:result:visual_SERV_CT}) present an even more demanding generalization test, as laparoscopic scenes differ substantially from colonoscopy in terms of illumination patterns, tissue appearance, and the presence of surgical instruments. 
Here, the predictions of EcoDepth degraded significantly, introducing spurious depth gradients along instrument shafts and failing to distinguish between overlapping anatomical structures. 
Marigold produced overly smoothed depth maps that collapse the critical depth separation between instruments and underlying tissue, potentially hazardous for surgical navigation. 
Depth Anything V2 maintained better scene coherence but struggled with the specular reflections common on moist tissue surfaces, creating localized depth artifacts. SpecDepth again demonstrated superior generalization, producing depth maps that maintain clear separation between instruments and tissue, preserve the curved geometry of abdominal organs, and handle specular reflections without introducing artifacts.

These external validation results provide compelling evidence that our spectral solution has learned physically-grounded depth cues which transcend dataset-specific characteristics. 
SpecDepth consistently preserved geometric boundaries across diverse endoscopic domains and handled specular reflections appropriately even on unseen tissue types, suggesting the model has internalized fundamental constraints of endoscopic image formation, including perspective geometry, illumination physics, and tissue reflectance properties. 
Unlike baseline methods, which exhibited dataset-specific shortcuts when transferred to new endoscopic contexts despite strong in-domain performance, SpecDepth maintained its robustness across domains. 
By demonstrating strong generalization to both polyp characterization and laparoscopic navigation tasks, our method positions itself as a foundation model capable of addressing physical constraints common to the broader endoscopic domain, with consistent performance across drastically different scene types, underscoring its potential as a versatile tool for endoscopic scene understanding.

\section{Discussion and Conclusion}

This study introduced SpecDepth, a parameter-efficient adaptation framework that addresses the fundamental challenge of transferring foundation models to colonoscopy depth estimation through spectral rectification rather than conventional semantic fine-tuning. 
Our key contribution lies in identifying that the performance degradation of natural-image pre-trained models on endoscopic data stems not from high-level semantic misalignment but from a statistical shift in the frequency domain: colonoscopy images exhibit severe high-frequency attenuation, lacking the strong edge and texture gradients that foundation models rely upon for geometric reasoning. 
To solve this spectral mismatch issue, we proposed an adaptive wavelet modulation mechanism that explicitly amplifies attenuated high-frequency components in feature maps, effectively realigning the input signal with the inductive bias of pre-trained models. 
Combined with a hybrid encoder that preserves shallow-layer representations while permitting deep-layer plasticity, and a multi-constraint objective that stabilizes spectral enhancement against noise amplification, SpecDepth achieved state-of-the-art performance on the C3VD and SimCol3D datasets. 
Quantitative results demonstrated substantial improvements over both general-purpose foundation models and task-specific depth estimators, with absolute relative errors of 0.022 and 0.027 on C3VD and SimCol3D, respectively, reducing error by approximately 79\% compared to the Depth Anything V2 baseline on C3VD. 
Ablation studies confirmed that each component contributed meaningfully to overall performance, with the wavelet gates learning to preferentially amplify high-frequency sub-bands during training. 
Qualitative visualizations further revealed that SpecDepth preserves crisp boundary delineation at fold structures while maintaining smooth, physically-plausible gradients in texture-sparse mucosal regions. 
External validation on colorectal polyp scenes and laparoscopic datasets demonstrated robust zero-shot generalization, confirming that spectral rectification addresses fundamental properties of endoscopic imagery rather than dataset-specific artifacts. 
Our work establishes that directly addressing spectral mismatches offers a highly effective and transferable strategy for adapting vision foundation models to specialized medical imaging domains.

Nonetheless, there are three limitations of this work.
First, the framework assumes static scene conditions during depth estimation, whereas real colonoscopy procedures involve dynamic tissue deformation, fluid motion, and instrument interaction that violate this assumption and can introduce temporal inconsistencies. 
Second, our SpecDepth utilizes a transformer-based backbone encoder, and the applicability of the spectral rectification paradigm to architectures with fundamentally different feature extraction mechanisms, including diffusion-based models and CNN encoders with skip-residual connections, remains to be established.
Third, although we validated generalization to laparoscopic scenes, the current evaluation was limited to qualitative assessment due to the absence of ground-truth depth in external datasets, making it difficult to quantify the precise extent of cross-domain transferability. 
Fourth, the spectral analysis revealed that colonoscopy images contain impulsive high-frequency components derived from mucosal artifacts and specular reflections, which our wavelet modulation may inadvertently amplify if not properly regularized. 
Finally, while parameter-efficient, the framework still requires supervised training with ground-truth depth, limiting its applicability in scenarios where such annotations are unavailable.

Future work will address these limitations through several directions. 
We plan to extend SpecDepth to handle dynamic scenes by incorporating temporal consistency constraints across video frames, enabling more robust depth estimation during instrument motion and tissue deformation. 
Another promising avenue is to develop self-supervised or weakly-supervised variants that can leverage unlabeled colonoscopy videos, reducing dependence on expensive ground-truth depth annotations. 
We also aim to investigate whether the spectral mismatch hypothesis extends to other modalities such as capsule endoscopy and bronchoscopy, where similar optical constraints operate, making spectral recalibration a more general component of foundation model adaptation in annotation-scarce medical imaging domains.
Furthermore, exploring alternative spectral decomposition techniques beyond wavelets, such as learnable filter banks or implicit neural representations, could provide more flexible control over frequency manipulation. 
Finally, we intend to validate SpecDepth on larger and more diverse endoscopic datasets, including prospective clinical studies, to establish its utility for real-time navigation, lesion localization, and computer-aided diagnosis. 
By continuing to bridge the spectral gap between natural and medical imaging domains, we believe this line of research will contribute to more reliable and generalizable foundation models for endoscopic scene understanding.


\bibliographystyle{model2-names}
\biboptions{authoryear}
\bibliography{A_refs}

@article{journal/MedIA/edwards2022,
  title={SERV-CT: A disparity dataset from cone-beam CT for validation of endoscopic 3D reconstruction},
  author={Edwards, PJ Eddie and Psychogyios, Dimitris and Speidel, Stefanie and Maier-Hein, Lena and Stoyanov, Danail},
  journal={Medical image analysis},
  volume={76},
  pages={102302},
  year={2022},
  publisher={Elsevier}
}

@article{journal/NC/ming2021,
  title={Deep learning for monocular depth estimation: A review},
  author={Ming, Yue and Meng, Xuyang and Fan, Chunxiao and Yu, Hui},
  journal={Neurocomputing},
  volume={438},
  pages={14--33},
  year={2021},
  publisher={Elsevier}
}

@article{journal/Medicine/Cianci2024,
  title={Colorectal cancer: prevention and early diagnosis},
  author={Cianci, Nicole and Cianci, Giulia and East, James E},
  journal={Medicine},
  volume={52},
  number={5},
  pages={251--257},
  year={2024},
  publisher={Elsevier}
}

@article{journal/MedIA/Bobrow2023,
  title={Colonoscopy 3D video dataset with paired depth from 2D-3D registration},
  author={Bobrow, Taylor L and Golhar, Mayank and Vijayan, Rohan and Akshintala, Venkata S and Garcia, Juan R and Durr, Nicholas J},
  journal={Medical image analysis},
  volume={90},
  pages={102956},
  year={2023},
  publisher={Elsevier}
}

@article{journal/TPAMI/Arampatzakis2023,
  title={Monocular depth estimation: A thorough review},
  author={Arampatzakis, Vasileios and Pavlidis, George and Mitianoudis, Nikolaos and Papamarkos, Nikos},
  journal={IEEE Transactions on Pattern Analysis and Machine Intelligence},
  volume={46},
  number={4},
  pages={2396--2414},
  year={2023},
  publisher={IEEE}
}

@inproceedings{conf/IPMI/Wang2023,
  title={A surface-normal based neural framework for colonoscopy reconstruction},
  author={Wang, Shuxian and Zhang, Yubo and McGill, Sarah K and Rosenman, Julian G and Frahm, Jan-Michael and Sengupta, Soumyadip and Pizer, Stephen M},
  booktitle={International Conference on Information Processing in Medical Imaging},
  pages={797--809},
  year={2023},
  organization={Springer}
}

@article{journal/Sensors/Masoumian2022,
  title={Monocular depth estimation using deep learning: A review},
  author={Masoumian, Armin and Rashwan, Hatem A and Cristiano, Juli{\'a}n and Asif, M Salman and Puig, Domenec},
  journal={Sensors},
  volume={22},
  number={14},
  pages={5353},
  year={2022},
  publisher={MDPI}
}

@article{oquab2023dinov2,
  title={Dinov2: Learning robust visual features without supervision},
  author={Oquab, Maxime and Darcet, Timoth{\'e}e and Moutakanni, Th{\'e}o and Vo, Huy and Szafraniec, Marc and Khalidov, Vasil and Fernandez, Pierre and Haziza, Daniel and Massa, Francisco and El-Nouby, Alaaeldin and Assran, Mahmoud and Ballas, Nicolas and Galuba, Wojciech and Howes, Russell and Huang, Po-Yao and Li, Shang-Wen and Misra, Ishan and Rabbat, Michael and Sharma, Vasu and Synnaeve, Gabriel and Xu, Hu and Jegou, Hervé and Mairal, Julien and Labatut, Patrick and Joulin, Armand and Bojanowski, Piotr},
  journal={arXiv preprint arXiv:2304.07193},
  year={2023}
}

@inproceedings{Geiger2012,
  title={Are we ready for autonomous driving? the kitti vision benchmark suite},
  author={Geiger, Andreas and Lenz, Philip and Urtasun, Raquel},
  booktitle={2012 IEEE conference on computer vision and pattern recognition},
  pages={3354--3361},
  year={2012},
  organization={IEEE}
}

@inproceedings{Silberman2012,
  title={Indoor segmentation and support inference from rgbd images},
  author={Silberman, Nathan and Hoiem, Derek and Kohli, Pushmeet and Fergus, Rob},
  booktitle={European conference on computer vision},
  pages={746--760},
  year={2012},
  organization={Springer}
}

@article{Wu2025,
  title={Medical sam adapter: Adapting segment anything model for medical image segmentation},
  author={Wu, Junde and Wang, Ziyue and Hong, Mingxuan and Ji, Wei and Fu, Huazhu and Xu, Yanwu and Xu, Min and Jin, Yueming},
  journal={Medical image analysis},
  volume={102},
  pages={103547},
  year={2025},
  publisher={Elsevier}
}

@article{Zhao2023,
  title={CLIP in medical imaging: A survey},
  author={Zhao, Zihao and Liu, Yuxiao and Wu, Han and Wang, Mei and Li, Yonghao and Wang, Sheng and Teng, Lin and Liu, Disheng and Cui, Zhiming and Wang, Qian and Shen, Dinggang},
  journal={Medical Image Analysis},
  volume={102},
  pages={103551},
  year={2025},
  publisher={Elsevier}
}

@article{journal/Access/Peng2024,
  title={PDLFBR-Net: Partial decoder localization and foreground-background refinement network for polyp segmentation},
  author={Peng, Yanbin and Feng, Mingkun and Zhai, Zhinian and Zheng, Zhijun},
  journal={IEEE Access},
  volume={12},
  pages={114280--114294},
  year={2024},
  publisher={IEEE}
}

@inproceedings{conf/MICCAI/Cui2024,
  title={Endodac: Efficient adapting foundation model for self-supervised depth estimation from any endoscopic camera},
  author={Cui, Beilei and Islam, Mobarakol and Bai, Long and Wang, An and Ren, Hongliang},
  booktitle={International Conference on Medical Image Computing and Computer-Assisted Intervention},
  pages={208--218},
  year={2024},
  organization={Springer}
}

@article{Zhang2023,
  title={Customized segment anything model for medical image segmentation},
  author={Zhang, Kaidong and Liu, Dong},
  journal={arXiv preprint arXiv:2304.13785},
  year={2023}
}

@article{Eigen2014,
  title={Depth map prediction from a single image using a multi-scale deep network},
  author={Eigen, David and Puhrsch, Christian and Fergus, Rob},
  journal={Advances in neural information processing systems},
  volume={27},
  year={2014}
}

@inproceedings{Godard2017b,
  title={Unsupervised monocular depth estimation with left-right consistency},
  author={Godard, Cl{\'e}ment and Mac Aodha, Oisin and Brostow, Gabriel J},
  booktitle={Proceedings of the IEEE conference on computer vision and pattern recognition},
  pages={270--279},
  year={2017}
}

@inproceedings{conf/ICCV/Godard2019,
  title={Digging into self-supervised monocular depth estimation},
  author={Godard, Cl{\'e}ment and Mac Aodha, Oisin and Firman, Michael and Brostow, Gabriel J},
  booktitle={Proceedings of the IEEE/CVF international conference on computer vision},
  pages={3828--3838},
  year={2019}
}

@inproceedings{conf/ICCV/Caron2021,
  title={Emerging properties in self-supervised vision transformers},
  author={Caron, Mathilde and Touvron, Hugo and Misra, Ishan and J{\'e}gou, Herv{\'e} and Mairal, Julien and Bojanowski, Piotr and Joulin, Armand},
  booktitle={Proceedings of the IEEE/CVF international conference on computer vision},
  pages={9650--9660},
  year={2021}
}

@inproceedings{conf/ICCV/Wang2023,
  title={Internimage: Exploring large-scale vision foundation models with deformable convolutions},
  author={Wang, Wenhai and Dai, Jifeng and Chen, Zhe and Huang, Zhenhang and Li, Zhiqi and Zhu, Xizhou and Hu, Xiaowei and Lu, Tong and Lu, Lewei and Li, Hongsheng and Wang, Xiaogang and Qiao, Yu},
  booktitle={Proceedings of the IEEE/CVF conference on computer vision and pattern recognition},
  pages={14408--14419},
  year={2023}
}

@inproceedings{Yang2024cvpr,
  title={Depth anything: Unleashing the power of large-scale unlabeled data},
  author={Yang, Lihe and Kang, Bingyi and Huang, Zilong and Xu, Xiaogang and Feng, Jiashi and Zhao, Hengshuang},
  booktitle={Proceedings of the IEEE/CVF conference on computer vision and pattern recognition},
  pages={10371--10381},
  year={2024}
}

@article{conf/NeurIPS/Yang2024,
  title={Depth anything v2},
  author={Yang, Lihe and Kang, Bingyi and Huang, Zilong and Zhao, Zhen and Xu, Xiaogang and Feng, Jiashi and Zhao, Hengshuang},
  journal={Advances in Neural Information Processing Systems},
  volume={37},
  pages={21875--21911},
  year={2024}
}

@article{journal/PR/Tian2022,
  title={Multi-stage image denoising with the wavelet transform},
  author={Tian, Chunwei and Zheng, Menghua and Zuo, Wangmeng and Zhang, Bob and Zhang, Yanning and Zhang, David},
  journal={Pattern Recognition},
  volume={134},
  pages={109050},
  year={2023},
  publisher={Elsevier}
}

@article{journal/pm/Lee2020,
  title={Sparse-view CT reconstruction based on multi-level wavelet convolution neural network},
  author={Lee, Minjae and Kim, Hyemi and Kim, Hee-Joung},
  journal={Physica Medica},
  volume={80},
  pages={352--362},
  year={2020},
  publisher={Elsevier}
}

@incollection{conf/MLAI/Sriwichai2022,
  title={On the Wavelet Convolution Neural Network for Breast Cancer Image Analysis},
  author={Sriwichai, Kittikorn and Onjun, Ratapong and Simtrakankul, Chantana and Kaennakham, Sayan},
  booktitle={Machine Learning and Artificial Intelligence},
  pages={105--110},
  year={2022},
  publisher={IOS Press}
}

@article{journal/MP/Qian2024,
  title={Adaptive wavelet-VNet for single-sample test time adaptation in medical image segmentation},
  author={Qian, Xiaoxue and Lu, Weiguo and Zhang, You},
  journal={Medical physics},
  volume={51},
  number={12},
  pages={8865--8881},
  year={2024},
  publisher={Wiley Online Library}
}

@article{journal/IEEE/Unser1996,
  title={A review of wavelets in biomedical applications},
  author={Unser, Michael and Aldroubi, Akram},
  journal={Proceedings of the IEEE},
  volume={84},
  number={4},
  pages={626--638},
  year={2002},
  publisher={IEEE}
}

@article{journal/SR/Yuan2024,
  title={Medical image segmentation with UNet-based multi-scale context fusion},
  author={Yuan, Yongqi and Cheng, Yong},
  journal={Scientific Reports},
  volume={14},
  number={1},
  pages={15687},
  year={2024},
  publisher={Nature Publishing Group UK London}
}

@article{journal/EIVP/Jin2015,
  title={Bilateral image denoising in the Laplacian subbands},
  author={Jin, Bora and You, Su Jeong and Cho, Nam Ik},
  journal={EURASIP Journal on Image and Video Processing},
  volume={2015},
  number={1},
  pages={26},
  year={2015},
  publisher={Springer}
}

@article{journal/RS/Tu2019,
  title={Hyperspectral image classification with multi-scale feature extraction},
  author={Tu, Bing and Li, Nanying and Fang, Leyuan and He, Danbing and Ghamisi, Pedram},
  journal={Remote sensing},
  volume={11},
  number={5},
  pages={534},
  year={2019},
  publisher={MDPI}
}

@article{journal/TNNLS/Liu2021,
  title={C-CNN: Contourlet convolutional neural networks},
  author={Liu, Mengkun and Jiao, Licheng and Liu, Xu and Li, Lingling and Liu, Fang and Yang, Shuyuan},
  journal={IEEE Transactions on Neural Networks and Learning Systems},
  volume={32},
  number={6},
  pages={2636--2649},
  year={2020},
  publisher={IEEE}
}

@article{journal/SP/Yuan2025,
  title={Multi-scale aggregation network for colonoscopic polyp segmentation via frequency domain decoupling},
  author={Wang, Yanling and Chin, Kho Lee and Song, Ngu Sze and Guo, Shengjin and Zhao, Wenjing},
  journal={Scientific Reports},
  volume={15},
  number={1},
  pages={44745},
  year={2025},
  publisher={Nature Publishing Group UK London}
}

@article{journal/BSPC/Wang2025,
  title={Medical image segmentation model based on multi-scale fusion and feature reconstruction convolution},
  author={Wang, Fuyao and Wang, Chuantao and Ma, Chi and Wang, Xiumin and Zhai, Jiliang and Zhao, Yu},
  journal={Biomedical Signal Processing and Control},
  volume={112},
  pages={108464},
  year={2026},
  publisher={Elsevier}
}

@article{journal/Frontiers/Zhang2025,
  title={Med-DGTN: Dynamic Graph Transformer with Adaptive Wavelet Fusion for multi-label medical image classification},
  author={Zhang, Guanyu and Li, Yan and Wang, Tingting and Shi, Guokun and Jin, Li and Gu, Zongyun},
  journal={Frontiers in Medicine},
  volume={12},
  pages={1600736},
  year={2025},
  publisher={Frontiers Media SA}
}

@inproceedings{conf/MICCAI/He2025,
  title={WDNet: A Novel Wavelet-Guided Hierarchical Diffusion Network for Multi-target Segmentation in Colonoscopy Images},
  author={He, Dongdong and Ma, Fang and Liu, Ziteng and Yin, Xunhai and Liu, Hao and Gao, Wenpeng and Zhang, Chenghong and Fu, Yili},
  booktitle={International Conference on Medical Image Computing and Computer-Assisted Intervention},
  pages={625--635},
  year={2025},
  organization={Springer}
}

@article{journal/PO/Tan2024,
  title={Colonoscopy polyp classification via enhanced scattering wavelet Convolutional Neural Network},
  author={Tan, Jun and Yuan, Jiamin and Fu, Xiaoyong and Bai, Yilin},
  journal={Plos one},
  volume={19},
  number={10},
  pages={e0302800},
  year={2024},
  publisher={Public Library of Science San Francisco, CA USA}
}

@article{conf/ICLR/Hu2022,
  title={Lora: Low-rank adaptation of large language models},
  author={Hu, Edward J and Shen, Yelong and Wallis, Phillip and Allen-Zhu, Zeyuan and Li, Yuanzhi and Wang, Shean and Wang, Liang and Chen, Weizhu},
  journal={Iclr},
  volume={1},
  number={2},
  pages={3},
  year={2022}
}

@article{Biderman2024,
  title={LoRA learns less and forgets less},
  author={Biderman, Dan and Portes, Jacob and Ortiz, Jose Javier Gonzalez and Paul, Mansheej and Greengard, Philip and Jennings, Connor and King, Daniel and Havens, Sam and Chiley, Vitaliy and Frankle, Jonathan and Blakeney, Cody and Cunningham, John P.},
  journal={arXiv preprint arXiv:2405.09673},
  year={2024}
}

@article{journal/TPAMI/Shao2024,
  title={NDDepth: Normal-distance assisted monocular depth estimation and completion},
  author={Shao, Shuwei and Pei, Zhongcai and Chen, Weihai and Chen, Peter CY and Li, Zhengguo},
  journal={IEEE Transactions on Pattern Analysis and Machine Intelligence},
  volume={46},
  number={12},
  pages={8883--8899},
  year={2024},
  publisher={IEEE}
}

@article{lee2019big,
  title={From big to small: Multi-scale local planar guidance for monocular depth estimation},
  author={Lee, Jin Han and Han, Myung-Kyu and Ko, Dong Wook and Suh, Il Hong},
  journal={arXiv preprint arXiv:1907.10326},
  year={2019}
}

@article{Solano2025,
  title={Multi-task learning with cross-task consistency for improved depth estimation in colonoscopy},
  author={Solano, Pedro Esteban Chavarrias and Bulpitt, Andrew and Subramanian, Venkataraman and Ali, Sharib},
  journal={Medical Image Analysis},
  volume={99},
  pages={103379},
  year={2025},
  publisher={Elsevier}
}

@article{wang2025monopcc,
  title={Monopcc: Photometric-invariant cycle constraint for monocular depth estimation of endoscopic images},
  author={Wang, Zhiwei and Zhou, Ying and He, Shiquan and Li, Ting and Huang, Fan and Ding, Qiang and Feng, Xinxia and Liu, Mei and Li, Qiang},
  journal={Medical Image Analysis},
  volume={102},
  pages={103534},
  year={2025},
  publisher={Elsevier}
}

@inproceedings{conf/CVPR/Bhat2021,
  title={Adabins: Depth estimation using adaptive bins},
  author={Bhat, Shariq Farooq and Alhashim, Ibraheem and Wonka, Peter},
  booktitle={Proceedings of the IEEE/CVF conference on computer vision and pattern recognition},
  pages={4009--4018},
  year={2021}
}

@article{journal/MedIA/Rau2024,
  title={SimCol3D—3D reconstruction during colonoscopy challenge},
  author={Rau, Anita and Bano, Sophia and Jin, Yueming and Azagra, Pablo and Morlana, Javier and Kader, Rawen and Sanderson, Edward and Matuszewski, Bogdan J and Lee, Jae Young and Lee, Dong-Jae and Posner, Erez and Frank, Netanel and Elangovan, Varshini and Raviteja, Sista and Li, Zhengwen and Liu, Jiquan and Lalithkumar, Seenivasan and Islam, Mobarakol and Ren, Hongliang and Lovat, Laurence B. and Stoyanov, Danail},
  journal={Medical Image Analysis},
  volume={96},
  pages={103195},
  year={2024},
  publisher={Elsevier}
}

@article{Leufkens2012,
  title={Factors influencing the miss rate of polyps in a back-to-back colonoscopy study},
  author={Leufkens, AM and Van Oijen, MGH and Vleggaar, FP and Siersema, PD},
  journal={Endoscopy},
  volume={44},
  number={05},
  pages={470--475},
  year={2012},
  publisher={Georg Thieme Verlag KG}
}

@article{xi2021,
  title={Global colorectal cancer burden in 2020 and projections to 2040},
  author={Xi, Yue and Xu, Pengfei},
  journal={Translational oncology},
  volume={14},
  number={10},
  pages={101174},
  year={2021},
  publisher={Elsevier}
}

@article{sung2021,
  title={Global cancer statistics 2020: GLOBOCAN estimates of incidence and mortality worldwide for 36 cancers in 185 countries},
  author={Sung, Hyuna and Ferlay, Jacques and Siegel, Rebecca L and Laversanne, Mathieu and Soerjomataram, Isabelle and Jemal, Ahmedin and Bray, Freddie},
  journal={CA: a cancer journal for clinicians},
  volume={71},
  number={3},
  pages={209--249},
  year={2021},
  publisher={Wiley Online Library}
}

@inproceedings{Martyniak2025,
  title={Simuscope: Realistic endoscopic synthetic dataset generation through surgical simulation and diffusion models},
  author={Martyniak, Sabina and Kaleta, Joanna and Dall'Alba, Diego and Naskret, Michal and Plotka, Szymon and Korzeniowski, Przemyslaw},
  booktitle={2025 IEEE/CVF Winter Conference on Applications of Computer Vision (WACV)},
  pages={4268--4278},
  year={2025},
  organization={IEEE}
}

@article{Jong2025,
  title={Impact of standard enhancement settings of endoscopy systems on performance of endoscopic artificial intelligence systems},
  author={Jong, Martijn R and Kusters, Carolus HJ and van Bokhorst, Querijn NE and Jukema, Jelmer B and van Heslinga, Rixta AH van Eijck and Fockens, Kiki N and Houwen, Britt BSL and Jaspers, Tim JM and Boers, Tim GW and van der Vlugt, Manon and Dekker, Evelien and Sommen, Fons van der and de, Peter H. N. and Groof, Albert J. de and Bergman, Jacques J. G. H. M},
  journal={Endoscopy},
  volume={57},
  number={06},
  pages={602--610},
  year={2025},
  publisher={Georg Thieme Verlag KG}
}

@inproceedings{conf/ECCVW/SheikhZeinoddin2025,
  title={Dares: Depth anything in robotic endoscopic surgery with self-supervised vector-lora of the foundation model},
  author={Sheikh Zeinoddin, Mona and Lena, Chiara and Qu, Jiongqi and Carlini, Luca and Magro, Mattia and Kim, Seunghoi and De Momi, Elena and Bano, Sophia and Grech-Sollars, Matthew and Mazomenos, Evangelos and Alexander, Daniel C. Alexander and Stoyanov, Danail and Clarkson, Matthew J. and Islam, Mobarakol},
  booktitle={European Conference on Computer Vision},
  pages={1--11},
  year={2024},
  organization={Springer}
}

@article{journal/IJMLC/Chen2021,
  title={Attention-based context aggregation network for monocular depth estimation},
  author={Chen, Yuru and Zhao, Haitao and Hu, Zhengwei and Peng, Jingchao},
  journal={International Journal of Machine Learning and Cybernetics},
  volume={12},
  number={6},
  pages={1583--1596},
  year={2021},
  publisher={Springer}
}

@inproceedings{conf/3DV/Laina2016,
  title={Deeper depth prediction with fully convolutional residual networks},
  author={Laina, Iro and Rupprecht, Christian and Belagiannis, Vasileios and Tombari, Federico and Navab, Nassir},
  booktitle={2016 Fourth international conference on 3D vision (3DV)},
  pages={239--248},
  year={2016},
  organization={IEEE}
}

@inproceedings{conf/CVPR/Zhou2017,
  title={Unsupervised learning of depth and ego-motion from video},
  author={Zhou, Tinghui and Brown, Matthew and Snavely, Noah and Lowe, David G},
  booktitle={Proceedings of the IEEE conference on computer vision and pattern recognition},
  pages={1851--1858},
  year={2017}
}

@article{journal/FrontRobAI/Mathew2023,
  title={Self-supervised monocular depth estimation for high field of view colonoscopy cameras},
  author={Mathew, Alwyn and Magerand, Ludovic and Trucco, Emanuele and Manfredi, Luigi},
  journal={Frontiers in Robotics and AI},
  volume={10},
  pages={1212525},
  year={2023},
  publisher={Frontiers Media SA}
}

@inproceedings{conf/BIBM/Zhou2023,
  title={Tackling challenges of low-texture and illumination variations for endoscopy self-supervised monocular depth estimation},
  author={Zhou, Luyan and Luo, Jingjing and Wang, Hongbo and Zhao, Shizun and Han, Yuan and Li, Wenxian},
  booktitle={2023 IEEE International Conference on Bioinformatics and Biomedicine (BIBM)},
  pages={2427--2432},
  year={2023},
  organization={IEEE}
}

@inproceedings{conf/MICCAI/Budd2024,
  title={Transferring relative monocular depth to surgical vision with temporal consistency},
  author={Budd, Charlie and Vercauteren, Tom},
  booktitle={International Conference on Medical Image Computing and Computer-Assisted Intervention},
  pages={692--702},
  year={2024},
  organization={Springer}
}

@inproceedings{conf/MICCAI/Wang2024,
  title={Structure-preserving image translation for depth estimation in colonoscopy},
  author={Wang, Shuxian and Paruchuri, Akshay and Zhang, Zhaoxi and McGill, Sarah and Sengupta, Roni},
  booktitle={International Conference on Medical Image Computing and Computer-Assisted Intervention},
  pages={667--677},
  year={2024},
  organization={Springer}
}

@article{journal/CIS/Zhao2023,
  title={WRANet: wavelet integrated residual attention U-Net network for medical image segmentation},
  author={Zhao, Yawu and Wang, Shudong and Zhang, Yulin and Qiao, Sibo and Zhang, Mufei},
  journal={Complex \& intelligent systems},
  volume={9},
  number={6},
  pages={6971--6983},
  year={2023},
  publisher={Springer}
}

@article{conf/BMVC/Hsieh2021,
  title={Learnable discrete wavelet pooling (LDW-Pooling) for convolutional networks},
  author={Wang, Bor-Shiun and Hsieh, Jun-Wei and Chang, Ming-Ching and Chen, Ping-Yang and Ke, Lipeng and Lyu, Siwei},
  journal={arXiv preprint arXiv:2109.06638},
  year={2021}
}

@inproceedings{conf/CVPR/Ramamonjisoa2021,
  title={Single image depth prediction with wavelet decomposition},
  author={Ramamonjisoa, Micha{\"e}l and Firman, Michael and Watson, Jamie and Lepetit, Vincent and Turmukhambetov, Daniyar},
  booktitle={Proceedings of the IEEE/CVF conference on computer vision and pattern recognition},
  pages={11089--11098},
  year={2021}
}

@article{journal/Sensors/Paul2023,
  title={Nested DWT--Based CNN Architecture for Monocular Depth Estimation},
  author={Paul, Sandip and Mishra, Deepak and Marimuthu, Senthil Kumar},
  journal={Sensors},
  volume={23},
  number={6},
  pages={3066},
  year={2023},
  publisher={MDPI}
}

@article{journal/Access/Liu2019,
  title={Multi-level wavelet convolutional neural networks},
  author={Liu, Pengju and Zhang, Hongzhi and Lian, Wei and Zuo, Wangmeng},
  journal={IEEE Access},
  volume={7},
  pages={74973--74985},
  year={2019},
  publisher={IEEE}
}

@article{journal/BOE/Wang2022,
  title={Wavelet attention network for the segmentation of layer structures on OCT images},
  author={Wang, Cong and Gan, Meng},
  journal={Biomedical optics express},
  volume={13},
  number={12},
  pages={6167--6181},
  year={2022},
  publisher={Optica Publishing Group}
}

@inproceedings{conf/MICCAI/Cheng2024,
  title={Winet: Wavelet-based incremental learning for efficient medical image registration},
  author={Cheng, Xinxing and Jia, Xi and Lu, Wenqi and Li, Qiufu and Shen, Linlin and Krull, Alexander and Duan, Jinming},
  booktitle={International Conference on Medical Image Computing and Computer-Assisted Intervention},
  pages={761--771},
  year={2024},
  organization={Springer}
}

@inproceedings{conf/IJCAI/Lu2025,
  title={Wavelet multi-scale region-enhanced network for medical image segmentation},
  author={Lu, Hang and Du, Liang and Zhou, Peng},
  booktitle={Proceedings of the Thirty-Fourth International Joint Conference on Artificial Intelligence},
  pages={1675--1683},
  year={2025}
}

@article{journal/CBM/Zhang2024,
  title={Segment anything model for medical image segmentation: Current applications and future directions},
  author={Zhang, Yichi and Shen, Zhenrong and Jiao, Rushi},
  journal={Computers in Biology and Medicine},
  volume={171},
  pages={108238},
  year={2024},
  publisher={Elsevier}
}

@article{journal/IJCARS/Cui2024,
  title={Surgical-dino: adapter learning of foundation models for depth estimation in endoscopic surgery},
  author={Cui, Beilei and Islam, Mobarakol and Bai, Long and Ren, Hongliang},
  journal={International Journal of Computer Assisted Radiology and Surgery},
  volume={19},
  number={6},
  pages={1013--1020},
  year={2024},
  publisher={Springer}
}

@article{gu2024build,
  title={How to build the best medical image segmentation algorithm using foundation models: a comprehensive empirical study with segment anything model. arXiv preprint arXiv: 240409957},
  author={Gu, H and Dong, H and Yang, J and Mazurowski, MA},
  journal={Cited Here},
  year={2024}
}

@article{conf/NIPS/Raghu2019,
  title={Transfusion: Understanding transfer learning for medical imaging},
  author={Raghu, Maithra and Zhang, Chiyuan and Kleinberg, Jon and Bengio, Samy},
  journal={Advances in neural information processing systems},
  volume={32},
  year={2019}
}

@inproceedings{conf/CVPR/Yuan2022,
  title={Neural window fully-connected crfs for monocular depth estimation},
  author={Yuan, Weihao and Gu, Xiaodong and Dai, Zuozhuo and Zhu, Siyu and Tan, Ping},
  booktitle={Proceedings of the IEEE/CVF conference on computer vision and pattern recognition},
  pages={3916--3925},
  year={2022}
}

@inproceedings{conf/CVPR/Patni2024,
  title={Ecodepth: Effective conditioning of diffusion models for monocular depth estimation},
  author={Patni, Suraj and Agarwal, Aradhye and Arora, Chetan},
  booktitle={Proceedings of the IEEE/CVF conference on computer vision and pattern recognition},
  pages={28285--28295},
  year={2024}
}

@inproceedings{conf/CVPR/ke2023,
  title={Repurposing diffusion-based image generators for monocular depth estimation},
  author={Ke, Bingxin and Obukhov, Anton and Huang, Shengyu and Metzger, Nando and Daudt, Rodrigo Caye and Schindler, Konrad},
  booktitle={Proceedings of the IEEE/CVF conference on computer vision and pattern recognition},
  pages={9492--9502},
  year={2024}
}

@inproceedings{conf/ECCV/shu2020,
  title={Feature-metric loss for self-supervised learning of depth and egomotion},
  author={Shu, Chang and Yu, Kun and Duan, Zhixiang and Yang, Kuiyuan},
  booktitle={European Conference on Computer Vision},
  pages={572--588},
  year={2020},
  organization={Springer}
}

@inproceedings{conf/AAAI/lyu2021,
  title={Hr-depth: High resolution self-supervised monocular depth estimation},
  author={Lyu, Xiaoyang and Liu, Liang and Wang, Mengmeng and Kong, Xin and Liu, Lina and Liu, Yong and Chen, Xinxin and Yuan, Yi},
  booktitle={Proceedings of the AAAI conference on artificial intelligence},
  volume={35},
  number={3},
  pages={2294--2301},
  year={2021}
}

@inproceedings{conf/BMVC/zhou2021,
    title={Self-Supervised Monocular Depth Estimation with Internal Feature Fusion},
    author={Zhou, Hang and Greenwood, David and Taylor, Sarah},
    booktitle={British Machine Vision Conference (BMVC)},
    year={2021}
}

@article{journal/MIA/shao2022,
  title={Self-supervised monocular depth and ego-motion estimation in endoscopy: Appearance flow to the rescue},
  author={Shao, Shuwei and Pei, Zhongcai and Chen, Weihai and Zhu, Wentao and Wu, Xingming and Sun, Dianmin and Zhang, Baochang},
  journal={Medical image analysis},
  volume={77},
  pages={102338},
  year={2022},
  publisher={Elsevier}
}

@article{journal/MIA/ozyoruk2021,
  title={EndoSLAM dataset and an unsupervised monocular visual odometry and depth estimation approach for endoscopic videos},
  author={Ozyoruk, Kutsev Bengisu and Gokceler, Guliz Irem and Bobrow, Taylor L and Coskun, Gulfize and Incetan, Kagan and Almalioglu, Yasin and Mahmood, Faisal and Curto, Eva and Perdigoto, Luis and Oliveira, Marina and Sahin, Hasan and Araujo, Helder and Alexandrino, Henrique and Durr Nicholas J and Gilbert, Hunter B and Turan, Mehmet},
  journal={Medical image analysis},
  volume={71},
  pages={102058},
  year={2021},
  publisher={Elsevier}
}

@inproceedings{conf/3DV/zhao2022,
  title={Monovit: Self-supervised monocular depth estimation with a vision transformer},
  author={Zhao, Chaoqiang and Zhang, Youmin and Poggi, Matteo and Tosi, Fabio and Guo, Xianda and Zhu, Zheng and Huang, Guan and Tang, Yang and Mattoccia, Stefano},
  booktitle={2022 international conference on 3D vision (3DV)},
  pages={668--678},
  year={2022},
  organization={IEEE}
}

@inproceedings{conf/CVPR/Zhang2023LiteMono,
  title={Lite-mono: A lightweight cnn and transformer architecture for self-supervised monocular depth estimation},
  author={Zhang, Ning and Nex, Francesco and Vosselman, George and Kerle, Norman},
  booktitle={Proceedings of the IEEE/CVF conference on computer vision and pattern recognition},
  pages={18537--18546},
  year={2023}
}

@article{journal/SD/song2025polyp,
  title={Polyp-size: a precise endoscopic dataset for AI-driven polyp sizing},
  author={Song, Yiming and Du, Sijia and Wang, Ruilan and Liu, Fei and Lin, Xiaolu and Chen, Jinnan and Li, Zeyu and Li, Zhao and Yang, Liuyi and Zhang, Zhengjie and Yan, Hao and Zhang, Qingwei and Qian, Dahong and Li, Xiaobo},
  journal={Scientific Data},
  volume={12},
  number={1},
  pages={918},
  year={2025},
  publisher={Nature Publishing Group UK London}
}

\end{document}